\documentclass[runningheads]{llncs}

\usepackage[mobile]{eccv}

\usepackage{eccvabbrv}

\usepackage{graphicx}
\usepackage{booktabs}
\usepackage{multirow}
\usepackage{wrapfig}
\usepackage[accsupp]{axessibility}  

\usepackage{hyperref}

\usepackage{orcidlink}

\usepackage[dvipsnames]{xcolor}
\usepackage[accsupp]{axessibility}

\usepackage{epsfig}
\usepackage{graphicx}
\usepackage{epstopdf}
\usepackage{amsmath}
\usepackage{amssymb}
\usepackage{color}
\usepackage{booktabs}
\usepackage{multirow}
\usepackage{url}
\usepackage[ruled, linesnumbered]{algorithm2e}
\usepackage{xcolor}
\usepackage{enumitem}

\usepackage{amsthm}
\usepackage{empheq}
\usepackage{latexsym}
\usepackage{array}
\usepackage{diagbox}
\usepackage{listings}
\usepackage[misc]{ifsym}
\usepackage[absolute]{textpos}

\definecolor{commentcolor}{rgb}{0,0,0}
\newcommand\blfootnote[1]{%
	\begingroup
	\renewcommand\thefootnote{}\footnote{#1}%
	\addtocounter{footnote}{-1}%
	\endgroup
}

\newenvironment{mytheorem}[1]{%
	\renewcommand\thetheorem{#1}
	\theorem
}{\endtheorem}

\def\E{\mathbf{E}}

\def\R{\mathbf{R}}
\def\M{\mathbf{M}}

\def\Q{\mathbf{Q}}

\def\x{\mathbf{x}}

\def\Q{\mathbf{Q}}

\DeclareMathOperator*{\quot}{quot}
\DeclareMathOperator*{\rank}{rank}

\begin{document}

\title{Six-Point Method for Multi-Camera Systems with Reduced Solution Space}

\definecolor{somegray}{gray}{0.5}
\newcommand{\darkgrayed}[1]{\textcolor{somegray}{#1}}
\begin{textblock}{11}(2.5, -0.1)  %
	\begin{center}
		\darkgrayed{This paper has been accepted for publication at the European Conference on Computer Vision (ECCV), 2024.}
	\end{center}
\end{textblock}

\titlerunning{Six-Point Method for Multi-Camera System}

\author{Banglei Guan\inst{1}$^\star$\orcidlink{0000-0003-2123-0182} \and
	Ji Zhao\inst{2}$^\star$\textsuperscript{(\Letter)}\orcidlink{0000-0002-0150-4601} \and
	Laurent Kneip\inst{3}\orcidlink{0000-0001-6727-6608}}

\authorrunning{B.~Guan, J. Zhao, and L. Kneip}

\institute{College of Aerospace Science and Engineering, National University \\ of Defense Technology, China. \email{guanbanglei12@nudt.edu.cn} \and
	Independent Researcher. Beijing, China. \email{zhaoji84@gmail.com} \blfootnote{$\star$ Equal contribution. \Letter~Corresponding author.} \\
	\and
	Mobile Perception Lab, ShanghaiTech University, China. \email{lkneip@shanghaitech.edu.cn}}

\maketitle

\begin{abstract}
Relative pose estimation using point correspondences (PC) is a widely used technique. A minimal configuration of six PCs is required for two views of generalized cameras. In this paper, we present several minimal solvers that use six PCs to compute the 6DOF relative pose of multi-camera systems, including a minimal solver for the generalized camera and two minimal solvers for the practical configuration of two-camera rigs. The equation construction is based on the decoupling of rotation and translation. Rotation is represented by Cayley or quaternion parametrization, and translation can be eliminated by using the hidden variable technique. Ray bundle constraints are found and proven when a subset of PCs relate the same cameras across two views. This is the key to reducing the number of solutions and generating numerically stable solvers. Moreover, all configurations of six-point problems for multi-camera systems are enumerated. Extensive experiments demonstrate the superior accuracy and efficiency of our solvers compared to state-of-the-art six-point methods. The code is available at \url{https://github.com/jizhaox/relpose-6pt}.

\end{abstract}

\begin{sloppypar}
\section{Introduction}
\label{sec:intro}
{R}{elative} pose estimation utilizing feature correspondences is a fundamental problem in geometric computer vision. It plays a crucial role in numerous tasks such as autonomous driving, augmented reality, simultaneous localization and mapping, etc. Despite having a long history, the research on relative pose estimation remains active. These efforts focus on enhancing the efficiency, stability, and accuracy of algorithms~\cite{guan2018visual,barath2020making,Eichhardt2020Relative,guanECCV2022,guan2023minimal}.

Camera models have a significant impact on computing the relative pose. A pinhole or perspective camera model is used to model a single camera~\cite{hartley2003multiple}, and more complicated cameras like multi-camera systems necessitate the use of a generalized camera model~\cite{grossberg2001general,sturm2004generic}. A generalized camera encapsulates various imaging systems by representing the landmark observations as spatial rays, which do not necessarily require originating from the projection center~\cite{pless2003using}. This paper is principally concerned with a multi-camera system comprising several cameras that have been installed rigidly. As we shall demonstrate in this paper, $n$ point correspondences (PCs) for a generalized camera can be represented similarly by using $2n$ single cameras in a multi-camera system. It is a proven fact that the standard epipolar geometry using five PCs is incapable of recovering the scale of translation~\cite{hartley2003multiple}. Conversely, the translation scale of multi-camera systems can be uniquely determined, and the minimum requirement for solving the relative pose increases from five to six PCs across both views~\cite{stewenius2005solutions}.

Due to the presence of outliers in PCs, a robust estimator is essential for accurate relative pose estimation and outlier rejection. The random sample consensus (RANSAC) framework~\cite{fischler1981random} and its various adaptations~\cite{lebeda2012fixing,raguram2012usac,barath2022graph,barath2022marginalizing} are widely employed in computer vision community. A minimal solver is the core component in the RANSAC framework. Using the epipolar geometry corresponding to each PC, a constraint can be derived to solve for the relative pose~\cite{hartley2003multiple}. Various methods for computing the relative pose of a single camera are known as five-point methods~\cite{nister2004efficient,stewenius2006recent,li2006five,kneip2012finding,Kukelova12polynomial,fathian2018quest}. The six-point method~\cite{stewenius2005solutions} is the first minimal solver proposed for computing the relative pose of a multi-camera system. Numerous methods have been introduced subsequently, such as the enhanced version of the six-point method~\cite{byrod2009fast}, the seventeen-point linear solvers~\cite{li2008linear,kim2009motion}, an iterative optimization-based solver~\cite{kneip2014efficient}, and a global optimization-based solver~\cite{zhao2020certifiably}.

This paper utilizes PCs to estimate the full DOF relative pose for multi-camera systems. The main contributions of this work are summarized as follows:
\begin{itemize}
	\item We estimate 6DOF relative pose from a minimal number of six PCs for multi-camera systems. By decoupling rotation and translation, a generic minimal solver for the generalized camera and two minimal solvers for popular configurations of two-camera rigs are proposed.
	\item For multi-camera systems, when a subset of PCs relate the same cameras across two views, ray bundle constraints are found and proven. It can be seen that using ray bundle constraints reduces the number of solutions and generates numerically stable solvers for relative pose estimation.
	\item All configurations of minimal six-point problems for multi-camera systems are first enumerated using graph enumeration and the P{\'o}lya enumeration theorem. Totally there are $5953$ cases. Moreover, we enumerate all the distinct graphs in a recursive way with 50 lines of Matlab code only. 
\end{itemize}

\section{Related Work}
The research on relative pose estimation remains active in geometric vision, with a lot of classical solvers existing in this area. First, these solvers can be divided into the relative pose estimation for single cameras~\cite{hartley2003multiple,nister2004efficient,stewenius2006recent,li2006five,kneip2012finding,Kukelova12polynomial,fathian2018quest} and generalized cameras~\cite{stewenius2005solutions,li2008linear,kim2009motion,kneip2014efficient,zhao2020certifiably}. These cameras can be calibrated or partially uncalibrated with unknown focal length or radial distortion.

Second, the relative pose estimation solvers can be categorized as minimal solvers~\cite{nister2004efficient,stewenius2005solutions}, non-minimal solvers~\cite{kim2009motion,kneip2013direct,kneip2014efficient,zhao2020certifiably,zhao2022efficient} and linear solvers~\cite{hartley1997defence,li2008linear}. The minimal solvers aim to estimate relative pose using the minimum number of geometric primitives. The non-minimal solvers leverage all the feature correspondences to compute the relative pose. The linear solvers demand a higher number of feature correspondences compared to minimal solvers and provide a straightforward and efficient solution. To achieve computational efficiency, the linear solvers often ignore implicit constraints on unknown parameters. Conversely, the minimal and non-minimal solvers usually exploit implicit constraints while solving for unknown parameters. When dealing with feature correspondences that may contain outliers, minimal solvers play a crucial role in ensuring robust estimation. For instance, RANSAC and its variants depend heavily on efficient minimal solvers.

The integration of these two aspects yields numerous subcategories. This paper specifically investigates minimal solvers for multi-camera systems using PCs. A minimal solver with $64$ solutions was first proposed to address the relative pose of multi-camera systems using $6$ PCs~\cite{stewenius2005solutions}. They also propose a special case with $56$ solutions when the extrinsic parameters satisfy a certain condition~\cite{stewenius2005solutions}. Then, a linear solver taking $17$ PCs was introduced in~\cite{li2008linear,kim2009motion}. Several solvers were proposed for applying to structure-from-motion with special configurations~\cite{zheng2015structure,kasten2019resultant}. Moreover, a few cases for semi-generalized cameras were proposed in~\cite{zheng2015structure}, which is a rather small subset of our enumerated cases. Several non-minimal solvers were proposed which leverage either local optimization~\cite{kneip2014efficient} or global optimization~\cite{zhao2020certifiably} to determine optimal relative poses. Some solvers required the motion priors of multi-camera systems, including known rotation axis~\cite{lee2014relative,sweeney2014solving,liu2017robust} and Ackermann motion~\cite{lee2013motion}. Moreover, an efficient solver was achieved by implementing a first-order approximation of relative rotation~\cite{ventura2015efficient}. By exploiting the additional affine parameters besides PCs, a minimal solver with two affine correspondences was proposed in~\cite{guanECCV2022}. Three significant differences between~\cite{guanECCV2022} and our method are clarified in the supplementary material.

\section{Relative Pose Estimation for Generalized Cameras}
\label{sec:pt_pose}
This section presents a novel six-point method for multi-camera systems and generalized cameras by decoupling rotation and translation. The ray bundle constraints are proven and exploited for solution space reduction. In addition, all configurations of minimal six-point problems for multi-camera systems are enumerated in this paper.

\subsection{Geometric Constraints}
\begin{figure}[tpb]
	\begin{center}
		\includegraphics[width=0.7\linewidth]{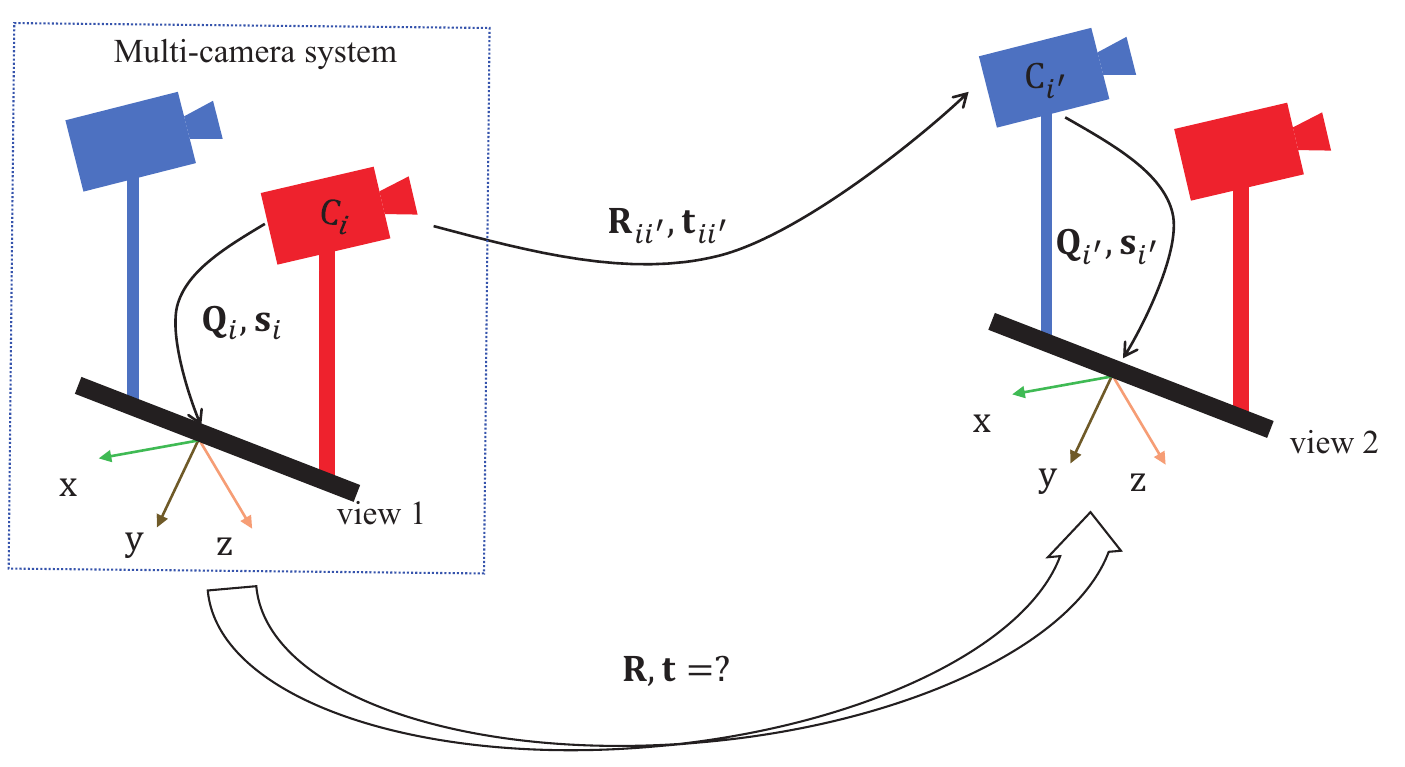}
	\end{center}
	\caption{Relative pose estimation for a multi-camera system. A point is observed by perspective camera $C_i$ in view 1 and by camera $C_{i'}$ in view 2. $\{\Q_i, \mathbf{s}_i\}$ and $\{\Q_{i'}, \mathbf{s}_{i'}\}$ are extrinsic parameters for $C_i$ and $C_{i'}$, respectively. The related point correspondence is described by two-view epipolar geometry of cameras $C_i$ and $C_{i'}$.}
	\label{fig:multi_cam_system_ac}
\end{figure}

\cref{fig:multi_cam_system_ac} illustrates a multi-camera system consisting of multiple perspective cameras, assuming that both the intrinsic and extrinsic parameters of these cameras have been calibrated. The extrinsic parameters of the camera $C_i$ are represented as $\{\Q_i, \mathbf{s}_i\}$, where $\Q_i$ represents the rotation and $\mathbf{s}_i$ represents the translation with respect to the multi-camera system's reference. Let $\R$ denote the rotation and $\mathbf{t}$ denote the translation between the first and second views.

A PC in a multi-camera system establishes the relationship of a point captured by two cameras across different views. Let the $k$-th PC be denoted by $(\x_k, \x'_k, i_k, i'_k)$. This indicates that the $i_k$-th camera observes a point in view~1, which is represented by its homogeneous coordinate $\x_k$ in the normalized image plane. Furthermore, this same point is also observed by the $i'_k$-th camera in view~2, which is represented by its homogeneous coordinate as $\x'_k$. For simplicity, we omit the subscript $k$ from camera indices $i$ and $i'$ to simplify the notation. In a multi-camera system, essential matrices vary for different PCs, distinguishing them from the two-view epipolar geometry of single cameras. Therefore, one constraint of epipolar geometry~\cite{hartley2003multiple} induced by the $k$-th PC is 
\begin{align}
	\x_{k'}^{T} \E_k \x_k = 0, \label{eq:essential_mat_pc}
\end{align}
where the essential matrix is represented as
\begin{align}
	\E_k = [\mathbf{t}_{ii'}]_\times \R_{ii'}.
	\label{eq:essential_gcam}
\end{align}
Here $\{\R_{ii'}, \mathbf{t}_{ii'}\}$ are the relative rotation and translation from camera $i$ in the first view to camera $i'$ in the second view. According to \cref{fig:multi_cam_system_ac}, it is obtained by a composition of three spatial transformations as 
\begin{align}
	\begin{bmatrix}
		\R_{ii'} & {\mathbf{t}_{ii'}}\\
		{{\mathbf{0}}}&{1}\\
	\end{bmatrix} 
	= &
	\begin{bmatrix}
		{\Q_{i'}}&{\mathbf{s}_{i'}}\\
		{{\mathbf{0}}}&{1}\\
	\end{bmatrix}^{-1}
	\begin{bmatrix}
		\R&{\mathbf{t}}\\
		{{\mathbf{0}}}&{1}\\
	\end{bmatrix}
	\begin{bmatrix}
		{\Q_{i}}&{\mathbf{s}_{i}}\\
		{{\mathbf{0}}}&{1}\\
	\end{bmatrix}  
	= & \begin{bmatrix} {\Q_{i'}^T \R \Q_i} & \Q_{i'}^T (\R \mathbf{s}_{i} + \mathbf{t} - \mathbf{s}_{i'})\\
		{{\mathbf{0}}}& \ {1}\\
	\end{bmatrix}.
	\label{eq:transformation}
\end{align}
By substituting $\R_{ii'}$ and $\mathbf{t}_{ii'}$ into Eq.~\eqref{eq:essential_gcam}, the essential matrix $\E_k$ can be reformulated as
\begin{align}
	\E_k = \Q_{i'}^T (\R [\mathbf{s}_i]_\times + [\mathbf{t} - \mathbf{s}_{i'}]_\times \R) \Q_i.
	\label{eq:essential_matrix}
\end{align}

Based on the above equation, it can be seen that Eq.~\eqref{eq:essential_mat_pc} is bilinear in the relative pose $\{\R, \mathbf{t}\}$.

\subsection{Relative Pose Parameterization}
We need to parametrize the relative pose of multi-camera systems. Rotation can be parameterized using Cayley parametrization, quaternions, Euler angles, direction cosine matrix (DCM), etc. Cayley and quaternion parameterizations have demonstrated superiority in minimal problems~\cite{zhao2020minimal}. Rotation matrix $\mathbf{R}$ using Cayley parameterization can be expressed as
\begin{align}
	\mathbf{R}_{\text{cayl}} = \small{\frac{1}{q_x^2+q_y^2+q_z^2+1}} {\begin{bmatrix}
			{1+q_x^2-q_y^2-q_z^2} &  2 q_x q_y -2 q_z & 2 q_x q_z + 2 q_y \\
			2 q_x q_y+2 q_z & 1-q_x^2+q_y^2-q_z^2 & 2 q_y q_z - 2 q_x \\
			2 q_x q_z - 2 q_y & 2 q_y q_z + 2 q_x & {1-q_x^2-q_y^2+q_z^2}
	\end{bmatrix}},	
	\label{eq:R6dof1}
\end{align}
where $[1,q_x,q_y,q_z]^T$ is a homogeneous quaternion vector. Rotation matrix $\R$ using quaternion parameterization can be written as
\begin{align}
	\mathbf{R}_{\text{quat}} = 	{\begin{bmatrix}
			q_w^2+q_x^2-q_y^2-q_z^2 &  2 q_x q_y -2 q_w q_z & 2 q_x q_z + 2 q_w q_y \\
			2 q_x q_y+2 q_w q_z & q_w^2-q_x^2+q_y^2-q_z^2 & 2 q_y q_z - 2 q_w q_x \\
			2 q_x q_z - 2 q_w q_y & 2 q_y q_z + 2 q_w q_x & q_w^2-q_x^2-q_y^2+q_z^2
	\end{bmatrix}},	
	\label{eq:R6dof2}
\end{align}
where $[q_w,q_x,q_y,q_z]^T$ is a quaternion vector satisfying the normalization constraint $q_w^2 + q_x^2 + q_y^2 + q_z^2 = 1$.

Note that $180^{\circ}$ rotations are not allowed in Cayley parameterization, although this is uncommon for typical image pairs. In practical applications, Cayley parameterization has been extensively utilized in minimal problems~\cite{stewenius2005minimal,kneip2014efficient,zheng2015structure,zhao2020minimal}. By contrast, quaternion parameterization does not have any degeneracy. However, quaternion introduces more variables than Cayley, so its solvers are usually less efficient. We construct solvers using both of these two parameterizations for completeness. In the following, we introduce the solver generation procedure based on Cayley parameterization. This approach can also be directly extended to quaternion parameterization.

The translation $\mathbf{t}$ can be parametrized as 
\begin{align}
	\mathbf{t} = \begin{bmatrix}
		{t_x}& \
		{t_y}& \
		{t_z}
	\end{bmatrix}^T.
	\label{eq:T6dof1}
\end{align} 

\subsection{Equation System Construction}
For multi-camera systems, the relative pose between two views is 6DOF. Thus, the relative pose estimation of a multi-camera system requires a minimal number of six PCs. Using Cayley parameterization, we obtain six polynomials for six unknowns $\{q_x, q_y, q_z, t_x, t_y, t_z\}$ from Eq.~\eqref{eq:essential_mat_pc} by substituting the essential matrix~\eqref{eq:essential_matrix} into them. After separating $q_x$, $q_y$, $q_z$ from $t_x$, $t_y$, $t_z$, we arrive at an equation system
\begin{align} 
	\underbrace {
		{\M}(q_x, q_y, q_z)}_{6\times 4}
	\begin{bmatrix}
		{{{t}_x}}\\
		{{{t}_y}}\\
		{{{t}_z}}\\
		1
	\end{bmatrix} = {\mathbf{0}},
	\label{eq:equ_qxqyqz1_pc}
\end{align}
where the entries of ${\M}$ are quadratic polynomials in three unknowns $q_x,q_y,q_z$. The $i$-th row corresponds to the constraint associated with the $i$-th PC. It can be observed that ${\M}$ has a null vector. Hence, the determinants of all the $4\times4$ submatrices of ${\M}$ must be zero.

Moreover, when a subset of PCs relates to the same perspective cameras across two views, there exists a specific property for this scenario. This property can be used to reduce the number of solutions and generate numerically stable solvers. Taking the number of a subset of PCs equal to 3 as an introductory example. When the number of a subset of PCs is greater than or equal to 3, it can be directly derived from Theorem~\ref{theorem:extra_constraint_pc}.

\begin{mytheorem}{1}
	\label{theorem:extra_constraint_pc}
	Denote $\mathcal{S}$ as a matrix set with elements satisfying $\mathbf{N} = {\M}([k_1,k_2,\\k_3], 1:3)$, where $\mathbf{N}$ is formed from rows $\{k_1,k_2,k_3\}$ and columns $\{1,2,3\}$ of $\mathbf{M}$. In addition, $k_1$-th, $k_2$-th, and $k_3$-th PCs are captured by the same perspective camera in each view and $k_1 < k_2 < k_3$. Then $\rank(\mathbf{N}) = 2$, $\forall \mathbf{N} \in \mathcal{S}$ holds for non-degenerate cases. 
\end{mytheorem}
\begin{proof}
	Let's proceed with the investigation of an arbitrary element in $\mathcal{S}$. We denote $\mathbf{N}_k$ as the $k$-th element in $\mathcal{S}$. The extrinsic parameters of the corresponding perspective camera in view~1 are denoted by $\{\Q_i, \mathbf{s}_i\}$, and for view~2, they are denoted by $\{\Q_{i^{\prime}}, \mathbf{s}_{i^{\prime}}\}$. 
	
	We begin by proving that $\rank(\mathbf{N}_k) \le 2$. To accomplish this objective, we must prove that the null space of $\mathbf{N}_k$ is non-empty. Given that the $k_1$-th, $k_2$-th, and $k_3$-th PCs are observed by the same perspective camera in each view, their associated essential matrices remain unchanged. Referring to Eq.~\eqref{eq:essential_matrix}, the essential matrix can be parametrized as
	\begin{align}
		\E_k = \Q_{i'}^T [\mathbf{t} + \R \mathbf{s}_i - \mathbf{s}_{i'}]_\times \R \Q_i.
	\end{align}
	
	Denote $\bar{\mathbf{t}} \triangleq \mathbf{t} + \R \mathbf{s}_i - \mathbf{s}_{i'}$, then we have
	\begin{align}
		\E_k = \Q_{i'}^T [\bar{\mathbf{t}}]_\times \R \Q_i.
		\label{eq:new_essential_matrix_app}
	\end{align}
	Substituting Eq.~\eqref{eq:new_essential_matrix_app} into Eq.~\eqref{eq:essential_mat_pc}, we obtain three equations for the three PCs. Each monomial in the three equations is linear with respect to one entry of vector $\bar{\mathbf{t}}$, and there is no constant term. Hence, these equations can be expressed as
	\begin{align}
		\frac{1}{q_x^2+q_y^2+q_z^2+1} \mathbf{A}_{k} \bar{\mathbf{t}} = \mathbf{0} \
		&\Rightarrow  \mathbf{A}_k (\mathbf{t} + \R \mathbf{s}_i - \mathbf{s}_{i'}) = \mathbf{0}, 
		\label{eq:null_space_app} \\
		&\Rightarrow
		\begin{bmatrix}
			\mathbf{A}_k & \mathbf{A}_k(\R \mathbf{s}_i - \mathbf{s}_{i'})
		\end{bmatrix}
		\begin{bmatrix}
			\mathbf{t} \\ 1
		\end{bmatrix}
		= \mathbf{0}.
		\label{eq:equ_qxqyqz1_new_form_app}
	\end{align}
	
	By comparing the construction procedure of Eq.~\eqref{eq:equ_qxqyqz1_pc} and Eq.~\eqref{eq:equ_qxqyqz1_new_form_app}, we can see that
	\begin{align}
		\mathbf{A}_k = {\M}([k_1,k_2,k_3], 1:3) = \mathbf{N}_k.
	\end{align}
	Substituting this equation into Eq.~\eqref{eq:null_space_app}, we can observe that the null space of $\mathbf{N}_k$ is indeed not empty. 
	
	Next we aim to prove that $\rank(\mathbf{N}_k)\ge2$. This goal can be achieved using proof by contradiction. If $\rank(\mathbf{N}_k)\le1$, then $\rank({\M}([k_1,k_2,k_3],{1:4}))\le2$ given that ${\M}([k_1,k_2,k_3],1:4)$ includes an extra column compared to $\mathbf{N}_k$. This degenerate case implies that the three PCs offer no more than two independent constraints for the relative pose estimation. This does not hold true for non-degenerate cases, hence invalidating the assumption that $\rank(\mathbf{N}_k)\le 1$. It can be seen that the rank of $\mathbf{N}$ is 2 in non-degenerate cases.
\end{proof}

We call the constraints in Theorem~\ref{theorem:extra_constraint_pc} as the \emph{ray bundle constraints}. In computer graphics, a ray bundle is a collection of light rays that share a common origin and propagate in different directions. In addition, a factor $ q_x^2+q_y^2+q_z^2+1$ can be factored out to simplify the equation system. Please see the supplementary material for proof. It leads to the generation of more efficient solvers while also potentially avoiding extraneous roots. In summary, the whole polynomial equation system consists of two types of constraints, which can be written as
\begin{align}
	{\mathcal{E}}_1 \triangleq \{ \quot(\det(\mathbf{N}), q_x^2+q_y^2+q_z^2+1) = 0 \ | \ 
	\mathbf{N} \in \small{4\times4 \text{ submatrices of }} {\M}\},
	\label{eq:submatrix_4by4_pc}
\end{align}
and
\begin{align}
	{\mathcal{E}}_2 \triangleq \{ \quot(\det(\mathbf{N}), q_x^2+q_y^2+q_z^2+1) = 0 \ | \ 
	\mathbf{N} \in \mathcal{S}\},
	\label{eq:submatrix_3by3_pc}
\end{align}
where $\quot(a, b)$ denotes the division quotient of $a$ by $b$, and $\det(\cdot)$ represents the determinant operator.

In ${\mathcal{E}}_1$, there are $15$ equations of degree $6$, while ${\mathcal{E}}_2$ contains some equations of degree 4. It should be noted that the number of equations in ${\mathcal{E}}_2$ varies depending on the specific configurations of the PCs. For certain PC configurations of multi-camera systems, ${\mathcal{E}}_2$ may be empty.    

After obtaining the rotation parameters $\{q_x, q_y, q_z \}$, the translation $[t_x, t_y, t_z]^T$ can be determined by initially calculating a vector within the null space of $\M$, followed by normalization where the last entry of the vector is divided.

\subsection{Polynomial System Solving} 
Based on the polynomial equation system Eqs.~\eqref{eq:submatrix_4by4_pc} and~\eqref{eq:submatrix_3by3_pc}, we propose a minimal generic solver for the generalized camera and two minimal solvers for standard configurations of two-camera rigs. The Gr{\"o}bner basis technique can be applied to discover algebraic solutions for the polynomial equation system~\cite{larsson2017efficient,martyushev2022optimizing}. Firstly, we construct a random instance of the original equation system in either a finite prime field $\mathbb{Z}_p$~\cite{lidl1997finite} or a rational field. This strategy helps to maintain numerical stability and avoid arithmetic with large numbers when computing Gr{\"o}bner basis. Secondly, \texttt{Macaulay~2}~\cite{grayson2002macaulay} is utilized for computing Gr{\"o}bner basis. Finally, we use an automatic Gr{\"o}bner basis solver~\cite{larsson2017efficient} to find the solution of the polynomial equation system. It should be noted that the polynomial equations $\mathcal{E}_1$ and $\mathcal{E}_2$ can be extended to deal with relative pose estimation for other configurations of multi-camera systems, such as with partially uncalibrated cameras and known rotation angles. 

\subsubsection{\label{sec:GenericSolver}Minimal Solver for Generalized Cameras}
\begin{figure}[tbp]
	\centering
	\begin{minipage}[t]{0.39\textwidth} 
		\centering
		\includegraphics[width=1.0\textwidth]{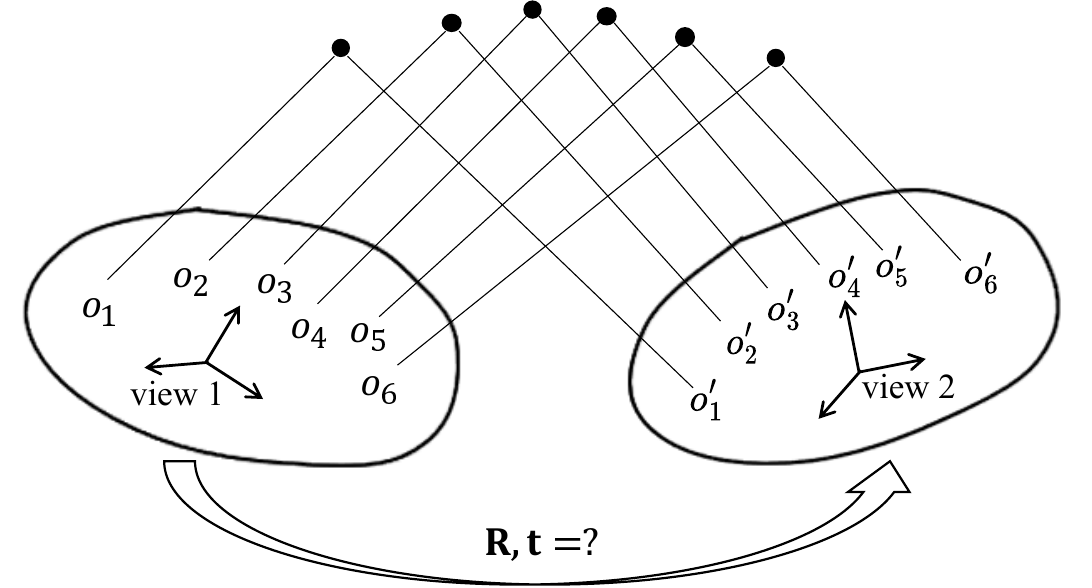}
		\caption{Relative pose estimation for generalized cameras. Note that points $o_i$ and $o'_i$ do not necessarily correspond to the same physical point of the generalized camera.}
		\label{fig:gcam}
	\end{minipage}
	\ \
	\begin{minipage}[t]{0.58\textwidth} 
		\centering
		\includegraphics[width=1.0\textwidth]{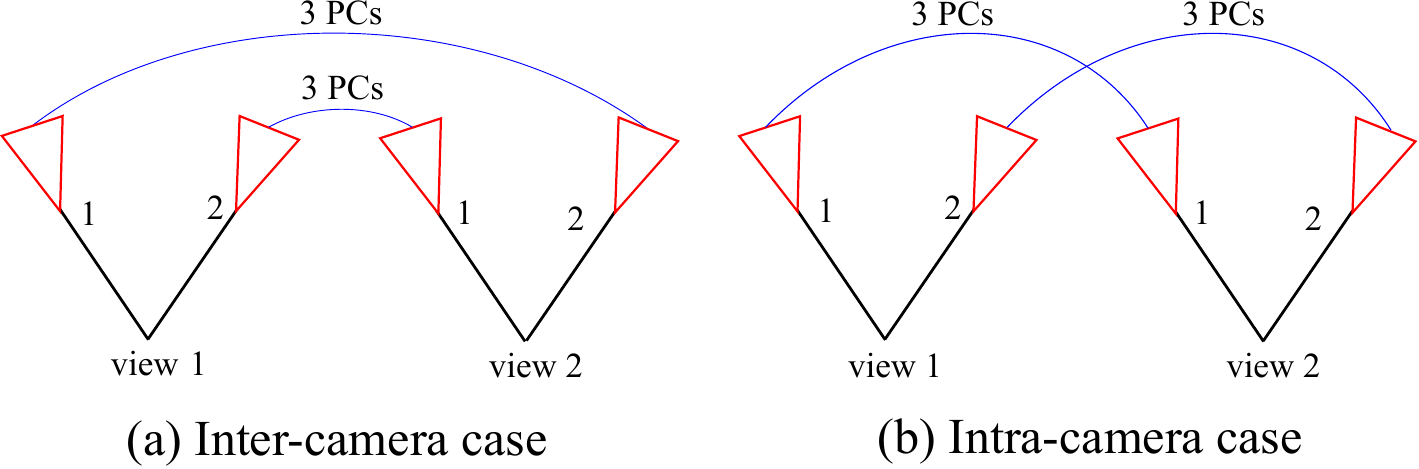}
		\caption{Relative pose estimation for two-camera rigs. Specifically, our goal is to determine the 6DOF relative pose while six PCs are observable by two views of a two-camera rig. (a) inter-camera case, (b) intra-camera case.}
		\label{fig:teaser_pc_pose}
	\end{minipage}
\end{figure}

We propose a generic solver for the generalized camera given six PCs. As shown in \cref{fig:gcam}, there are two views of a generalized camera, and there are 6 PCs across two views. Specifically, two-view geometry for 6 PCs and a generalized camera can be described by a 12-camera rig. We can define 12 virtual perspective cameras by the following method. The origins $o_1, \cdots, o_6$ and $o'_1, \cdots, o'_6$ of the virtual cameras are the positions of PCs. The orientations of these virtual cameras are consistent with the generalized camera's reference. The 6 PCs can be equivalently captured by a virtual 12-camera rig. Moreover, we can calculate the extrinsic parameters of these 12 virtual perspective cameras and the image coordinates of PCs in these virtual cameras. Please see the supplementary material for details. As a result, we can construct a minimal generic solver to recover the relative pose of generalized cameras. 

\cref{tab:complete_solution_pc} shows the statistics of the proposed minimal solvers for generalized cameras. Here, \texttt{\#sym} indicates the number of symmetries, \texttt{\#sol} indicates the number of solutions, and \texttt{$1$-dim} indicates one-dimensional extraneous roots. The Cayley parameterization and quaternion parameterization solvers are named as \texttt{6pt+cayl+generic} and \texttt{6pt+quat+generic}, respectively. 
The observations are summarized as follows: (1) $\mathcal{E}_2$ is an empty set, and $\mathcal{E}_1$ is sufficient to solve the relative pose. (2) Due to one-fold symmetry in quaternion, Cayley parameterization results in fewer solutions than quaternion parameterization. The number of complex solutions yielded by the \texttt{6pt+cayl+generic} solver and \texttt{6pt+quat+generic} solver is 64 and 128, respectively. (3) Elimination templates of the \texttt{6pt+cayl+generic} solver and the \texttt{6pt+quat+generic} solver are $99\times 163$ and $342\times406$, respectively. Given that the solvers using Cayley parameterization yields smaller eliminate templates compared to the solvers using quaternion parameterization, we adopt the former as our default choice in this paper.

Moreover, we enumerate all configurations of minimal six-point problems for multi-camera systems. The P{\'o}lya enumeration theorem can be applied to solve this problem~\cite{guichard2023combinatorics}, and a combinatorics solution shows that there are $5953$ cases totally in this problem. Most of these cases can be solved by the generic solver. Please see the supplementary material for details.
\begin{table}[tbp]
	\centering
	\caption{Minimal solvers for relative pose estimation. cayl: Cayley parameterization; quat: quaternion parameterization; inter: inter-camera PCs; intra: intra-camera PCs. For the generic case, $\mathcal{E}_2$ is an empty set and ${\mathcal{E}}_1 \cup {\mathcal{E}}_2 = \mathcal{E}_1$.}
	\label{tab:complete_solution_pc}
	\setlength{\tabcolsep}{2pt}{
		\scalebox{0.92}{
			\begin{tabular}{l c c c c c c} 
				\hline
				\multirow{2}{*}{\centering configuration} &  \multicolumn{3}{c}{equations ${\mathcal{E}}_1$} &  \multicolumn{3}{c}{equations ${\mathcal{E}}_1 \cup {\mathcal{E}}_2$}  \\ 
				\cline{2-7} 
				&   \#sym &  \#sol  & template   &  \#sym &  \#sol    &   template  \\ 
				\hline
				6pt+cayl+generic & $0$ & $64$ & $99\times 163$ & $0$ & $64$ & $99\times 163$  \\
				6pt+cayl+inter & 0 & $56$ & $56\times 120$ & 0 & $48$ & $64\times 120$  \\ 
				6pt+cayl+intra & 0 & $1$-dim & $-$ & 0 & $48$ & $72\times 120$  \\ \hline  
				6pt+quat+generic & 1 & $128$ & $342\times 406$  & 1 & $128$ &  $342\times 406$ \\
				6pt+quat+inter & 1 & $112$ & $174\times 243$  & 1 & $96$ &  $152\times 200$ \\ 
				6pt+quat+intra & 1 & $1$-dim & $-$ & 1 & $96$ & $152\times 200$  \\ \hline
			\end{tabular}
		}
	}
\end{table}

\subsubsection{\label{sec:InterIntraSolver}Minimal Solvers for Two-camera Rigs} 
Two minimal solvers are proposed for two practical configurations of two-camera rigs in \cref{fig:teaser_pc_pose}. These solvers comprise an inter-camera solver and an intra-camera solver, and both configurations offer two ray bundle constraints within $\mathcal{E}_2$. The inter-camera solver utilizes inter-camera PCs that are observable to different cameras across two views. This solver is appropriate for multi-camera systems characterized by significant overlap between views. Conversely, the intra-camera solver utilizes intra-camera PCs that are observable to the same camera across two views. This solver is appropriate for multi-camera systems characterized by small or no overlap between views.

For inter-camera case, $\mathcal{E}_1$ is enough to compute the relative pose of two-camera rigs. The number of solutions can be reduced by employing both $\mathcal{E}_1$ and $\mathcal{E}_2$. For intra-camera case, one-dimensional families of extraneous roots exist when only $\mathcal{E}_1$ is employed. Combining $\mathcal{E}_1$ and $\mathcal{E}_2$ can solve the relative pose in the intra-camera case. The solvers for these two cases using Cayley parametrization are named as \texttt{6pt+cayl+inter} and \texttt{6pt+cayl+intra}, respectively. The solvers for these two cases using quaternion parameterization are named as \texttt{6pt+quat+inter} and \texttt{6pt+quat+intra}, respectively.

\cref{tab:complete_solution_pc} also shows the statistics of our minimal solvers for two-camera rigs.
The observations are summarized as follows: (1) When $\mathcal{E}_1$ is employed, the inter-camera solver has up to $56$ complex solutions, and there exist one-dimensional families of extraneous roots in the intra-camera case. (2) By employing both $\mathcal{E}_1$ and $\mathcal{E}_2$, both the inter-camera and intra-camera solvers yield a total of 48 complex solutions. (3) Compared to the solvers using quaternion parameterization, the solvers using Cayley parameterization yield smaller eliminate templates. We adopt Cayley parameterization as our default choice. (4) When only $\mathcal{E}_1$ is used in the \texttt{6pt+quat+inter} configuration, it is necessary to explicitly consider the inequality $q_w \neq 0$. Otherwise, one-dimensional extraneous roots exist. We account for this inequality utilizing the saturation method~\cite{larsson2017polynomial}, which yields $112$ solutions exhibiting one-fold symmetry. (5) For the \texttt{6pt+cayl+inter} configuration, compared to the solvers derived from both $\mathcal{E}_1$ and $\mathcal{E}_2$, the solvers solely derived from $\mathcal{E}_1$ yield smaller eliminate templates and exhibit better numerical stability. This phenomenon indicates that the number of bases might impact the numerical stability of the solvers, which has been previously observed in the literature~\cite{byrod2009fast,guanECCV2022}. 

\section{Experiments} 
\label{sec:experiment}
This section presents a series of experiments performed on synthetic and real-world datasets to assess the performance of our solvers. All the proposed solvers employ Cayley parameterization for their implementations. The solver for the generalized camera is named as the \texttt{6pt-Our-generic} method. The solvers aim to handle inter-camera and intra-camera cases, are named as the \texttt{6pt-Our-inter} and \texttt{6pt-Our-intra} solvers, respectively. To further differentiate between different solvers for \texttt{6pt-Our-inter}, we denote the solvers resulting from ${\mathcal{E}}_1$ and ${\mathcal{E}}_1 \cup {\mathcal{E}}_2$ as \texttt{6pt-Our-inter56} and \texttt{6pt-Our-inter48}, respectively. Following~\cite{kneip2014efficient,guanECCV2022}, our solvers are evaluated against state-of-the-art solvers using PCs, including \texttt{17pt-Li}~\cite{li2008linear}, \texttt{8pt-Kneip}~\cite{kneip2014efficient}, and \texttt{6pt-Stew{\'e}nius}~\cite{stewenius2005solutions}. The paper does not evaluate solvers that exploit the additional affine parameters besides PCs~\cite{Guan_ICRA2021,guanECCV2022} or utilize a prior for relative pose estimation~\cite{sweeney2014solving,ventura2015efficient}. The proposed minimal solvers are implemented in C++. The source codes for \texttt{17pt-Li}~\cite{li2008linear} and \texttt{8pt-Kneip}~\cite{kneip2014efficient} are adopted from the OpenGV library~\cite{kneip2014opengv}. The source codes for \texttt{6pt-Stew{\'e}nius}~\cite{stewenius2005solutions} is adopted from the PoseLib library~\cite{PoseLib}.

In the experiments presented in \cref{sec:syn} and \cref{sec:real}, each solver is independently integrated into RANSAC~\cite{fischler1981random} to reject outliers and obtain the estimated relative pose with the highest number of inliers. To ensure the fairness of the experiment, all the different solvers are compared over the same data and within the same RANSAC implementation. In addition, we do not apply a non-minimal solver or perform optimization for the estimated relative pose with all inliers. The angular re-projection error~\cite{kneip2014opengv,lee2019closed} induced by PCs is used to classify inliers. We follow the default parameters of OpenGV to set the inlier threshold angle as $0.1^\circ$~\cite{kneip2014opengv}. 
During RANSAC iterations, the outlier ratio is from the current best model. The stopping criterion of RANSAC iterations is that at least one outlier-free set is sampled with a probability $0.99$, or the maximum number $20,000$ of iterations is reached. We compute the rotation error as the angular difference between the ground truth rotation $\mathbf{R}_{\text{gt}}$ and the estimated rotation ${\mathbf{R}}$: ${\varepsilon_{\mathbf{R}}} = \arccos ((\mathrm{tr}({\mathbf{R}_{\text{gt}}}{{\mathbf{R}^T}}) - 1)/2)$. To calculate the translation error, we employ the definition introduced in~\cite{quan1999linear,lee2014relative}: ${\varepsilon_{\mathbf{t}}} = 2\left\| {{\mathbf{t}_{\text{gt}}}}-{\mathbf{t}}\right\|/(\left\| {\mathbf{t}_{\text{gt}}} \right\| + \left\| {{\mathbf{t}}} \right\|)$, where $\mathbf{t}_{\text{gt}}$ and ${\mathbf{t}}$ are the ground truth translation and the estimated translation, respectively. We also compute the translation direction error: $\varepsilon_{\mathbf{t},\text{dir}} = \arccos(\mathbf{t}_{\text{gt}}^T \mathbf{t} / (\|\mathbf{t}_{\text{gt}}\| \cdot \|\mathbf{t}\|))$.

The principles of minimal solver choice are listed below. First, when \texttt{6pt-Our-inter} or \texttt{6pt-Our-intra} are applicable, we apply them with higher priority than \texttt{6pt-Our-generic}. The reason is that the solvers of specific configurations usually have better performance than the generic solver. Second, we recommend \texttt{6pt-Our-intra} as the default solver for real-world image sequences. The relative pose is usually estimated for consecutive image pairs captured in a small time interval. Thus, intra-camera PCs can be built up with a higher probability than inter-camera PCs. The results for efficiency and numerical stability are shown in the supplementary material.

\subsection{Experiments on Synthetic Data}
\label{sec:syn}
Two simulated scenarios are designed and tested for the synthetic experiments. First, we design a simulated two-camera rig composed of two perspective cameras. The orientations of the two perspective cameras are roughly forward-facing with random perturbation. This setting is practical for autonomous driving with two front-facing cameras. Second, we design a simulated generalized camera composed of 12 omnidirectional cameras. The extrinsic parameters, including orientation and position, are totally random. The results for two-camera rigs under the first scenario are shown in the supplementary material.
\begin{figure*}[tbp]
	\centering
	\includegraphics[width=0.50\linewidth]{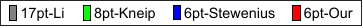}\\
	\centering
	\begin{subfigure}{0.32\linewidth}
		\includegraphics[width=1.0\linewidth]{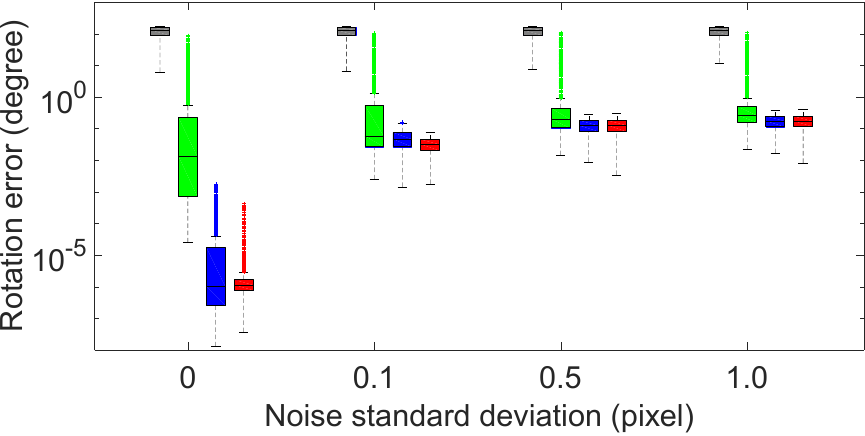}
		\caption{${\varepsilon_{\mathbf{R}}}$ in generic case.}
	\end{subfigure}
	\hfill
	\begin{subfigure}{0.32\linewidth}
		\includegraphics[width=1.0\linewidth]{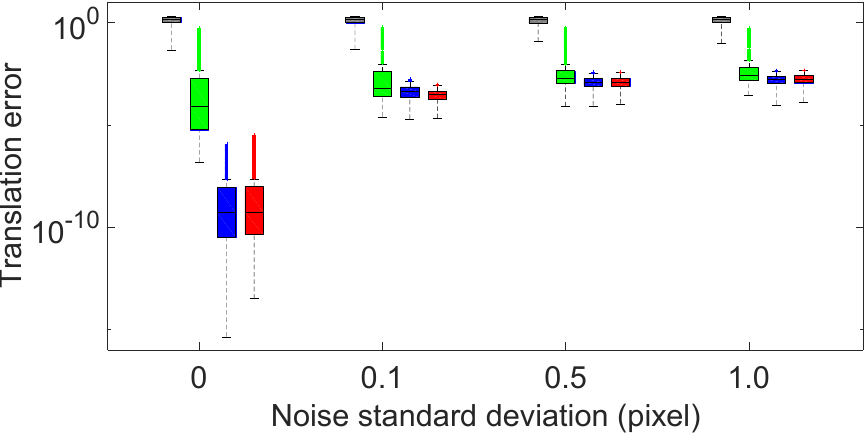}
		\caption{$\varepsilon_{\mathbf{t}}$ in generic case.}
	\end{subfigure}
	\hfill
	\begin{subfigure}{0.32\linewidth}
		\includegraphics[width=1.0\linewidth]{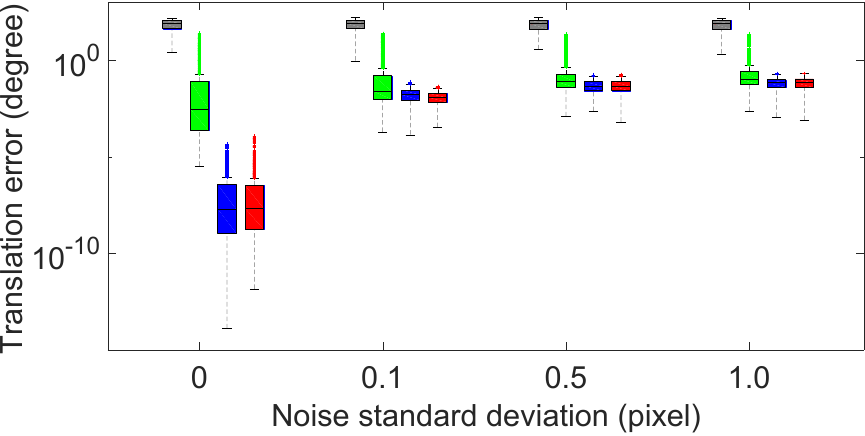}
		\caption{$\varepsilon_{\mathbf{t},\text{dir}}$ in generic case.}
	\end{subfigure}
	\begin{subfigure}{0.32\linewidth}
		\includegraphics[width=1.0\linewidth]{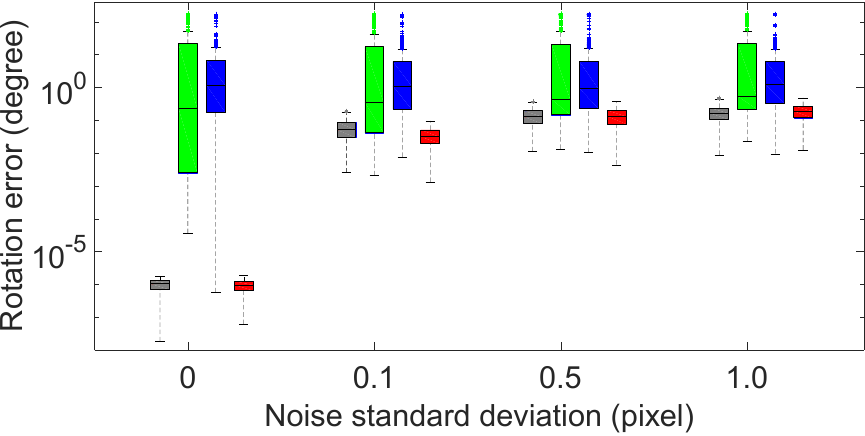}
		\caption{${\varepsilon_{\mathbf{R}}}$ in inter-camera case.}
	\end{subfigure}
	\hfill
	\begin{subfigure}{0.32\linewidth}
		\includegraphics[width=1.0\linewidth]{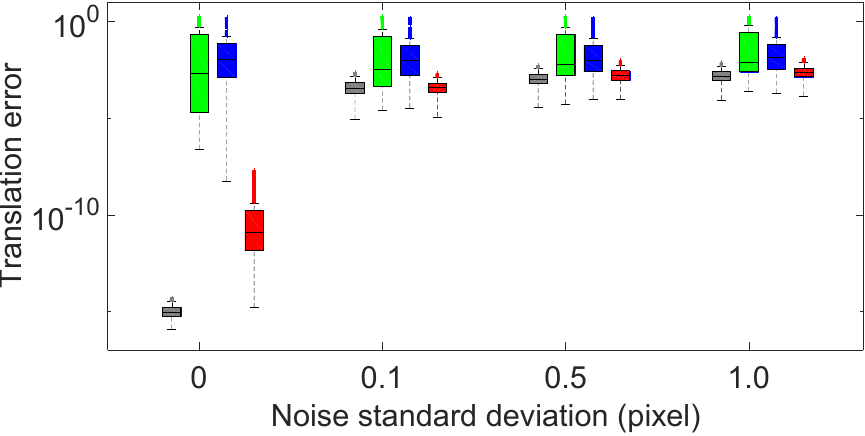}
		\caption{$\varepsilon_{\mathbf{t}}$ in inter-camera case.}
	\end{subfigure}
	\hfill
	\begin{subfigure}{0.32\linewidth}
		\includegraphics[width=1.0\linewidth]{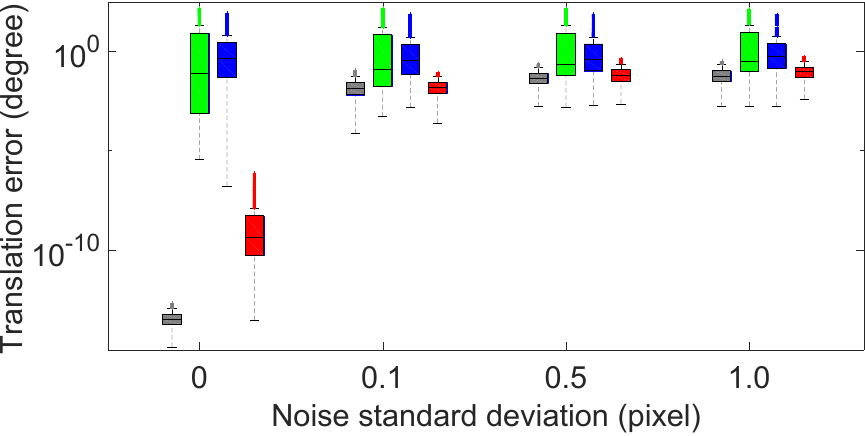}
		\caption{$\varepsilon_{\mathbf{t},\text{dir}}$ in inter-camera case.}
	\end{subfigure}
	\begin{subfigure}{0.32\linewidth}
		\includegraphics[width=1.0\linewidth]{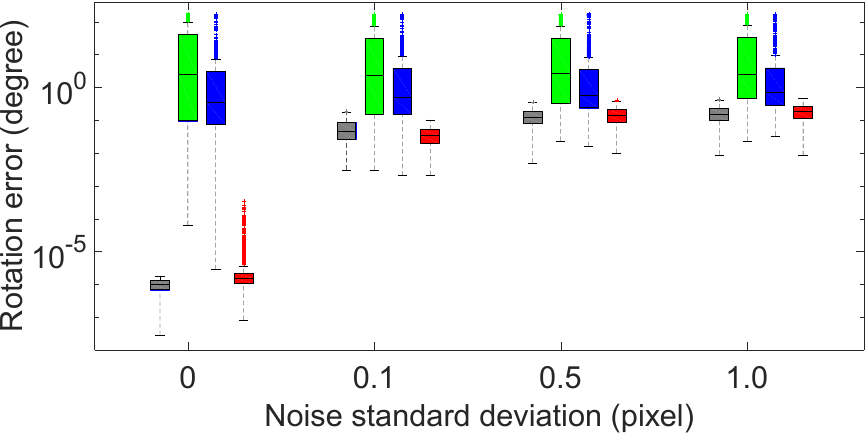}
		\caption{${\varepsilon_{\mathbf{R}}}$ in intra-camera case.}
	\end{subfigure}
	\hfill
	\begin{subfigure}{0.32\linewidth}
		\includegraphics[width=1.0\linewidth]{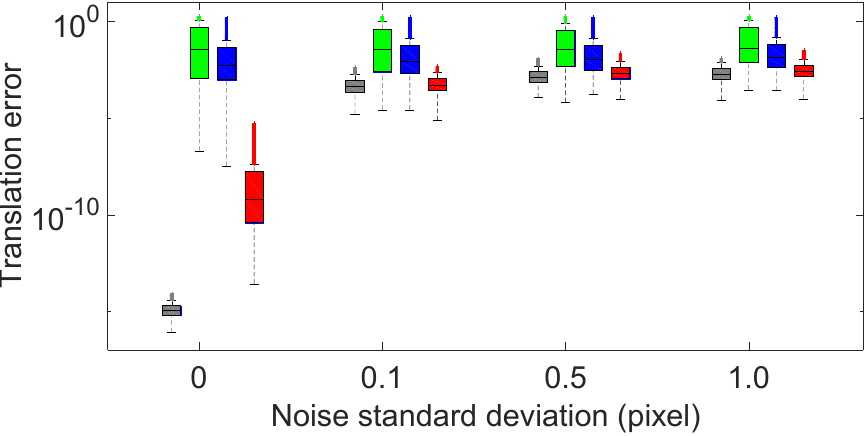}
		\caption{$\varepsilon_{\mathbf{t}}$ in intra-camera case.}
	\end{subfigure}
	\hfill
	\begin{subfigure}{0.32\linewidth}
		\includegraphics[width=1.0\linewidth]{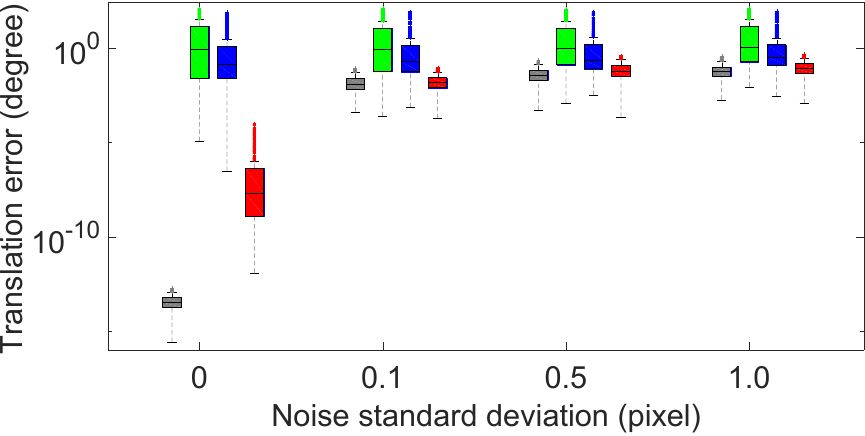}
		\caption{$\varepsilon_{\mathbf{t},\text{dir}}$ in intra-camera case.}
	\end{subfigure}
	\caption{Relative pose estimation error with varying image noise for a generalized camera. We design a simulated multi-camera system comprising 12 omnidirectional cameras. The extrinsic parameters, including orientation and position, are totally random. The three rows correspond to the generic, inter-camera, and intra-camera cases, respectively. The \texttt{6pt-Our} solver in the three rows represent \texttt{6pt-Our-generic}, \texttt{6pt-Our-inter}, and \texttt{6pt-Our-intra}, respectively.}
	\label{fig:generic_inter_intra}
\end{figure*}

Under the second scenario, the simulated scenario for a generalized camera is described as below. For the generic case, a generalized camera comprises 12 omnidirectional cameras. For the inter-camera and intra-camera cases, we only use 2 omnidirectional cameras of the generalized camera. The extrinsic parameters of each omnidirectional camera are generated randomly. Each scene point is generated randomly. In addition, the relative poses between two views are also generated randomly. Omnidirectional cameras are selected in our settings, because usually there is no overlap for pinhole cameras with random extrinsic parameters and relative poses. We test the accuracy of pose estimation for all the proposed solvers, including \texttt{6pt-Our-generic}, \texttt{6pt-Our-inter}, and \texttt{6pt-Our-intra} solvers.

Some of the comparison solvers use more PCs than the proposed solvers. The strategy of correspondence selection has a significant influence on their performance. We design a rule to select matches for different cases to guarantee fairness. Please see the supplementary material for details. In the synthetic experiments, we conduct $1000$ trials for each case and a specific noise level combined with the RANSAC framework. For each trial, we randomly generate 100 PCs and select correspondences randomly for the different solvers. \cref{fig:generic_inter_intra} illustrates the performance of various solvers against image noise for generalized cameras. The observations are summarized as follows: (1) \texttt{17pt-Li} has good overall accuracy for inter-camera and intra-camera cases. However, it fails for the generic case due to rank deficiency, and the essential matrix with scale ambiguity cannot be uniquely recovered. (2) \texttt{8pt-Kneip} has acceptable results according to the median metric. However, the error variance is large for all the cases. (3) \texttt{6pt-Stew{\'e}nius} has good overall accuracy for the generic case. It does not work well for inter-camera and intra-camera cases, and the error variance is large for both cases. This phenomenon is consistent with~\cite{kim2009motion}, which observes this solver does not work for most axial cameras where every bearing vector intersects a line in 3D. (4) The proposed \texttt{6pt-Our} solver works for all the cases and has satisfactory overall accuracy.

\subsection{Experiments on Real-World Data}
\label{sec:real}
To assess the performance of our solvers in practical applications, three datasets \texttt{KITTI}~\cite{geiger2013vision}, \texttt{nuScenes}~\cite{Caesar_2020_CVPR}, and \texttt{EuRoc}~\cite{burri2016euroc} are used in the experiments. Specifically, the \texttt{KITTI} and \texttt{nuScenes} datasets are collected in an autonomous driving environments, while the \texttt{EuRoc} dataset is collected in a micro aerial vehicle environment. These datasets contain challenging image pairs with highly dynamic scenes, such as significant motion, moving objects, and varying illumination. The proposed solvers are compared against state-of-the-art solvers. The accuracy of relative pose estimation is evaluated using the rotation error ${\varepsilon _{\bf{R}}}$ and the translation direction error $\varepsilon_{\mathbf{t},\text{dir}}$~\cite{kneip2014efficient,liu2017robust}. Our evaluation is based on approximately $30,000$ image pairs, and we report the final estimation results by integrating the minimal solver with RANSAC. The relative pose estimation results for the \texttt{nuScenes} dataset are provided in the supplementary material.
\begin{table}[htbp]
	\caption{Rotation and translation error on \texttt{KITTI} dataset (unit: degree).}
	\begin{center}
		\setlength{\tabcolsep}{1.0mm}{
			\scalebox{0.890}{
				\begin{tabular}{c||c|c|c|c}
					\hline
					\multirow{2}{*}{{Seq.}} &  
					{17pt-Li}\small{~\cite{li2008linear}} &  {8pt-Kneip}\small{~\cite{kneip2014efficient}} &  {6pt-Stew.}\small{~\cite{stewenius2005solutions}}& {6pt-Our-intra} \\
					\cline{2-5}
					& ${\varepsilon _{\bf{R}}}$\quad\ $\varepsilon_{\mathbf{t},\text{dir}}$      &  ${\varepsilon _{\bf{R}}}$\quad\ $\varepsilon_{\mathbf{t},\text{dir}}$      &   ${\varepsilon _{\bf{R}}}$\quad\ $\varepsilon_{\mathbf{t},\text{dir}}$     &   ${\varepsilon _{\bf{R}}}$\quad\ $\varepsilon_{\mathbf{t},\text{dir}}$  \\
					\hline
					{00}&       0.147  \  2.537&    0.148  \  2.496&	0.243  \  4.521    &\textbf{0.136} \ \textbf{2.415}     \\
					{01}&       0.178  \  4.407&	0.182  \  3.485&	0.293  \  7.187    &\textbf{0.175} \ \textbf{3.323}     \\
					{02}&       0.142  \  1.988&	0.147  \  2.094&	0.227  \  3.315    &\textbf{0.139} \ \textbf{1.897}     \\
					{03}&\textbf{0.126}  \  2.762&	0.139  \  2.833&	0.314  \  6.254    &0.143          \ \textbf{2.740}     \\
					{04}&       0.113  \  1.733&	0.123  \  1.829&	0.262  \  3.670    &\textbf{0.101} \ \textbf{1.677}     \\
					{05}&       0.132  \  2.663&	0.130  \  2.461&	0.216  \  4.212    &\textbf{0.128} \ \textbf{2.342}     \\
					{06}&       0.139  \  2.146&	0.151  \  2.145&	0.197  \  3.240    &\textbf{0.121} \ \textbf{2.064}     \\
					{07}&       0.131  \  3.085&	0.172  \  3.259&	0.259  \  6.664    &\textbf{0.129} \ \textbf{2.904}     \\
					{08}&\textbf{0.133} \ 2.705&	0.135  \  2.762&	0.217  \  4.590    & 0.140         \ \textbf{2.620}     \\
					{09}&0.144  \  2.022&	0.138  \ \textbf{1.974}&    0.210  \  3.204    &\textbf{0.126} \  2.002     \\
					{10}&       0.142  \  2.398&	0.141  \  2.393&	0.246  \  3.849    &\textbf{0.137} \ \textbf{2.314}     \\
					\hline					
		\end{tabular}}}
	\end{center}
	\label{tab:RTErrror_kitti_generalized}
\end{table}

\begin{table}[htbp]
	\caption{Average runtime of RANSAC on \texttt{KITTI} dataset (unit:~$s$).}
	\begin{center}
		\setlength{\tabcolsep}{1.0mm}{
			\scalebox{0.839}{
				\begin{tabular}{c||c|c|c|c}
					\hline
					{Methods} &  {17pt-Li}\small{~\cite{li2008linear}} &  {8pt-Kneip}\small{~\cite{kneip2014efficient}} &  {6pt-Stew.}\small{~\cite{stewenius2005solutions}}& {6pt-Our-intra} \\
					\hline
					{Mean time }    & 3.157 &   0.648  &  4.161& 1.546\\
					\hline
					{Std. deviation}& 0.119 &   0.009  &  0.145& 0.069\\
					\hline
		\end{tabular}}}
	\end{center}
	\label{RANSACTime_generalized}
\end{table}

\subsubsection{\label{sec:KITTIexperiments}Experiments on KITTI Dataset}
We evaluate the proposed solvers on \texttt{KITTI} dataset~\cite{geiger2013vision}, which is collected using outdoor autonomous vehicles installed with forward-facing stereo cameras. It is treated as a general multi-camera system, disregarding overlapping overlapping fields of view for cameras. The \texttt{6pt-Our-intra} solver is tested on 11 available sequences containing 23,000 image pairs. The ground truth is obtained directly from the GPS/IMU localization unit~\cite{geiger2013vision}. To establish PCs for consecutive views of each camera, the SIFT method~\cite{lowe2004distinctive} is used. Additionally, all the solvers have been integrated into RANSAC to remove mismatches in the experiments.

\cref{tab:RTErrror_kitti_generalized} illustrates the rotation and translation error of the \texttt{6pt-Our-intra} solver on the \texttt{KITTI} dataset. We use median error to evaluate the performance of solvers. The proposed \texttt{6pt-Our-intra} solver outperforms the comparative solvers in overall performance. Since \texttt{8pt-Kneip} uses an identity matrix to initialize the rotation, it has acceptable results for the forward motion in autonomous driving environment. To further compare computational efficiency, \cref{RANSACTime_generalized} illustrates the corresponding average runtime of RANSAC on the \texttt{KITTI} dataset. Although the runtime of \texttt{17pt-Li} is lower than the proposed \texttt{6pt-Our-intra}, the proposed solver demonstrates better efficiency when each is integrated separately into the RANSAC framework.

\subsubsection{\label{sec:EuRoCexperiments}Experiments on EuRoC Dataset}
The proposed solvers are further validated within the context of an unmanned aerial vehicle (UAV) environment, leveraging the \texttt{EuRoC} MAV dataset~\cite{burri2016euroc} for the evaluation of 6DOF relative pose estimation. This dataset records data using a stereo camera mounted on a micro aerial vehicle, with sequences labeled from MH01 to MH05 collected within a large industrial machine hall. The ground truth for the relative pose is established through a nonlinear least-squares batch solution utilizing Leica position and IMU measurements. Estimating relative pose estimation in these sequences is challenging due to the unstructured and cluttered nature of the industrial environment. Consecutive image pairs with small movement are selectively thinned out, retaining only one out of every four consecutive images for relative pose estimation. Moreover, the image pairs exhibiting insufficient motion are excluded from the experiment. The proposed solvers are evaluated against state-of-the-art solvers, including \texttt{17pt-Li}~\cite{li2008linear}, \texttt{8pt-Kneip}~\cite{kneip2014efficient}, and \texttt{6pt-Stew{\'e}nius}~\cite{stewenius2005solutions}. PCs in the image pair are established using the SIFT detector~\cite{lowe2004distinctive}. All solvers are integrated into the RANSAC framework to eliminate mismatches, ensuring a more robust estimation process.
\begin{table}[tbp]
	\caption{Rotation and translation error on \texttt{EuRoC} dataset (unit: degree).}
	\begin{center}
		\setlength{\tabcolsep}{1.0mm}{
			\scalebox{0.93}{
				\begin{tabular}{c||c|c|c|c}
					\hline
					\multirow{2}{*}{{Seq.}} &  
					{17pt-Li}\small{~\cite{li2008linear}} &  {8pt-Kneip}\small{~\cite{kneip2014efficient}} &  {6pt-Stew.}\small{~\cite{stewenius2005solutions}}& \ {6pt-Our-intra} \\
					\cline{2-5}
					& ${\varepsilon _{\bf{R}}}$\quad\ $\varepsilon_{\mathbf{t},\text{dir}}$      &  ${\varepsilon _{\bf{R}}}$\quad\ $\varepsilon_{\mathbf{t},\text{dir}}$      &   ${\varepsilon _{\bf{R}}}$\quad\ $\varepsilon_{\mathbf{t},\text{dir}}$     &   ${\varepsilon _{\bf{R}}}$\quad\ $\varepsilon_{\mathbf{t},\text{dir}}$ \\
					\hline
					{MH01}& 0.136  \  3.055&	0.156  \  3.214&	0.186  \  4.085 &\textbf{0.130} \ \textbf{2.961}  \\
					{MH02}& 0.129  \  2.806&	0.132  \  2.796&	0.180  \  3.828 &\textbf{0.127} \ \textbf{2.579}  \\
					{MH03}& 0.199  \  2.422&	0.187  \  2.517&	0.222  \  3.576 &\textbf{0.181} \ \textbf{2.376}  \\
					{MH04}& 0.195  \  3.159&	0.178  \  3.237&	0.213  \  5.371 &\textbf{0.193} \ \textbf{3.105}  \\
					{MH05}& 0.186  \  3.124&	0.163  \  2.940&	0.235  \  4.094 &\textbf{0.158} \ \textbf{2.892}  \\
					\hline													
		\end{tabular}}}
	\end{center}
	\label{tab:RTErrror_EuRoC_generalized}
\end{table}	

\cref{tab:RTErrror_EuRoC_generalized} illustrates the rotation and translation error of the \texttt{6pt-Our-intra} solver on \texttt{EuRoC} dataset. We use median error to assess the performance of different solvers. The experiment results demonstrate that the \texttt{6pt-Our-intra} solver surpasses comparative solvers, including \texttt{17pt-Li}, \texttt{8pt-Kneip}, and \texttt{6pt-Stew{\'e}nius}. This experiment confirms the suitability of the proposed \texttt{6pt-Our-intra} solver for achieving accurate 6DOF relative pose estimation in the context of unmanned aerial vehicle environments.

\section{\label{sec:conclusion}Conclusion} 
We proposed a series of minimal solvers to compute the 6DOF relative pose of multi-camera systems using a minimal number of six PCs. We also exploit ray bundle constraints that allow for a reduction of the solution space and the development of more stable solvers. A generic solver is proposed for the relative pose estimation of generalized cameras. All configurations of minimal six-point problems for multi-camera systems are enumerated. Moreover, two minimal solvers, including an inter-camera solver and an intra-camera solver, are proposed for practical configurations of two-camera rigs. Based on both synthetic and real-world experiments, we demonstrate that the proposed solvers offer an efficient solution for estimating ego-motion of multi-camera systems, surpassing state-of-the-art solvers in terms of accuracy.

\section*{Acknowledgments}
This research has been supported in part by the Hunan Provincial Natural Science Foundation for Excellent Young Scholars under Grant 2023JJ20045, and the National Natural Science Foundation of China under Grant 12372189. Further funding support is provided by project 62250610225 by the Natural Science Foundation of China, as well as projects 22DZ1201900, 22ZR1441300, and dfycbj-1 by the Natural Science Foundation of Shanghai. J. Zhao would like to thank Saibal Mitra, from the Netherlands, for the valuable discussions on graph enumeration.

\end{sloppypar}
%
%
\bibliographystyle{splncs04}
\bibliography{main}

\newpage
\appendix


\title{Supplementary Material for\\ Six-Point Method for Multi-Camera Systems with Reduced Solution Space} 

\titlerunning{Six-Point Method for Multi-Camera System}

\author{Banglei Guan\inst{1}$^\star$\orcidlink{0000-0003-2123-0182} \and
Ji Zhao\inst{2}$^\star$\textsuperscript{(\Letter)}\orcidlink{0000-0002-0150-4601} \and
Laurent Kneip\inst{3}\orcidlink{0000-0001-6727-6608}}

\authorrunning{B.~Guan, J. Zhao, and L. Kneip}

\institute{College of Aerospace Science and Engineering, National University \\ of Defense Technology, China. \email{guanbanglei12@nudt.edu.cn} \and
Independent Researcher. Beijing, China. \email{zhaoji84@gmail.com} \blfootnote{$\star$ Equal contribution. \Letter~Corresponding author.} \\
\and
Mobile Perception Lab, ShanghaiTech University, China. \email{lkneip@shanghaitech.edu.cn}}

\maketitle

\appendix
\section*{Appendix}

\begin{sloppypar}

\section{Related Work}
Given that each affine correspondence yields three geometric constraints, a minimal solver using two affine correspondences was proposed to estimate the 6DOF relative pose~\cite{guanECCV2022}, which is related to our method in this paper. Although both methods use the classical hidden variable technique to decouple rotation and the other variables, and use Gr{\"o}bner basis technique to find algebraic solutions, we clarify that there are three significant differences as follows: (1) Different parametrizations of the relative pose problem are used, which leads to different polynomial equation systems. A further fundamental difference lies in the geometric constraints derived from point or affine correspondences. Even if the highest-degree monomial is the same, the polynomial equation systems are still different. (2) Finding practical solvers for the six-point problem using the rotation-translation parametrization is certainly non-trivial and there are no off-the-shelf solvers. This requires instantiating random problems in a finite prime field while preserving all geometric side constraints. (3) Two contributions are newly proposed. The ray bundle constraints are proven and exploited for solution space reduction. The ray bundle shares a common origin and propagates in different directions, which is different from the implicit geometric constraints in~\cite{guanECCV2022}. Moreover, all configurations of minimal six-point problems for multi-camera systems are first enumerated using the P{\'o}lya enumeration theorem. 

\section{Proof of a Common Factor}
In Section~3.3 of the paper, we say that a factor $q_x^2+q_y^2+q_z^2+1$ can be factored out in Eqs.~(14) and (15) of the paper. The existence of a common factor can be proven through formula derivation~\cite{zhao2021relative}. In this paper, we show that this property can be also proven by symbolic computation using Matlab. The Matlab code is listed in \cref{lst:proof}.
{
\renewcommand{\thelstlisting}{\arabic{lstlisting}}
\scriptsize
\begin{lstlisting}[caption=Proof by Symbolic Computation using Matlab, label={lst:proof}]
%% define an orthogonal matrix via Cayley's formula
syms x y z real
Q = [1+x^2-y^2-z^2, 2*x*y-2*z, 2*y+2*x*z; 2*x*y+2*z, 1-x^2+y^2-z^2, 2*y*z-2*x; 2*x*z-2*y, 2*x+2*y*z, 1-x^2-y^2+z^2];
		
%% check orthogonality
% The results are s^2*diag([1, 1, 1]),
% where s = x^2+y^2+z^2+1.
simplify(Q*Q')
		
%% construct a 3*3 random matrix N in Eq.(15)
imax = 10;
c = cell(3, 1);
for i = 1:3
  c{i} = cross(randi(imax, [3,1]), Q*randi(imax, [3,1]));
end
C1 = [c{1}, c{2}, c{3}];
eq1 = det(C1);
		
%% check the determinant has factor x^2+y^2+z^2+1
factor(eq1)
		
%% construct a 4*4 random matrix N in Eq.(14)
c = cell(4, 1);
mm = [x^2; y^2; z^2; x*y; x*z; y*z; x; y; z; 1];
for i = 1:4
  tmp = randi(imax, [1, 10])*mm;
  c{i} = [cross(randi(imax, [3,1]), Q*randi(imax, [3,1])); tmp];
end
C2 = [c{1}, c{2}, c{3}, c{4}];
eq2 = det(C2);
		
%% check the determinant has factor x^2+y^2+z^2+1
factor(eq2)
\end{lstlisting}
}

It can be seen that $3\times 3$ and $4\times 4$ submatrices of ${\M}(q_x, q_y, q_z)$ in Eq.~(8) of the paper have a common factor $q_x^2+q_y^2+q_z^2+1$. Note that representing all variables with symbols, that is, changing random numbers in the code to symbols, makes a rigorous proof. However, the runtime of symbolic computation will become about $15$ hours. 

Based on the existence of the common factor, we can use this property to factor out $q_x^2+q_y^2+q_z^2+1$ which simplifies the equation system. It leads to the generation of more efficient solvers while also potentially avoiding false roots.

\section{Six-Point Solvers for Generalized Cameras}
In Section~3.4 of the paper, we discuss the application of the six-point method to the generalized camera model, which is a concept larger than multiple camera systems. Specifically, we show that six-point matches of a generalized camera can be equivalently captured by virtual 12-perspective-camera rigs, making the six-point method applicable to any generalized camera.
	\begin{figure}[htbp]
		\centering
		\begin{subfigure}{0.58\linewidth}
			\includegraphics[width=1.0\linewidth]{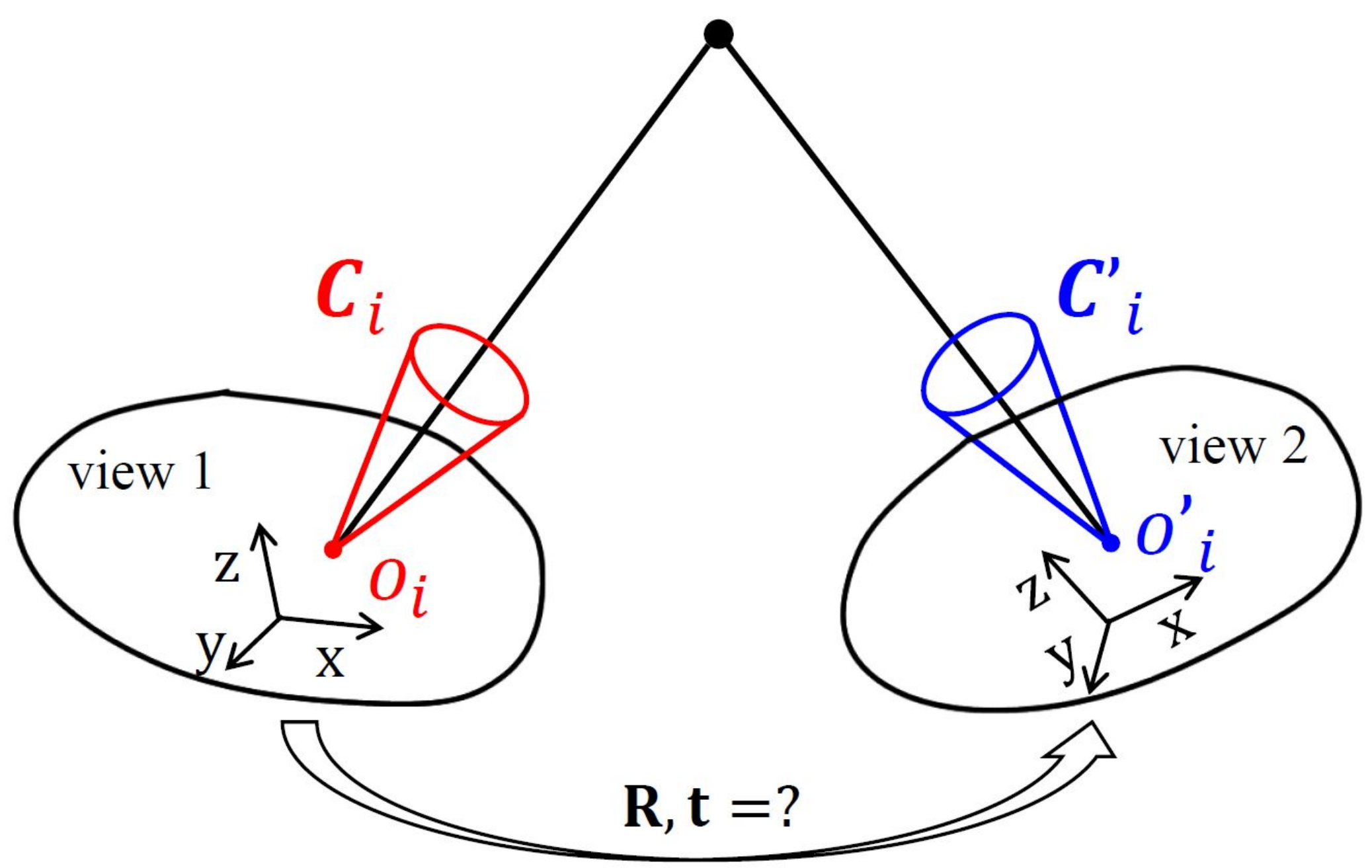}
			\caption{$i$-th match of a generalized camera.}
		\end{subfigure}
		\ \ \ \
		\begin{subfigure}{0.368\linewidth}
			\includegraphics[width=1.0\linewidth]{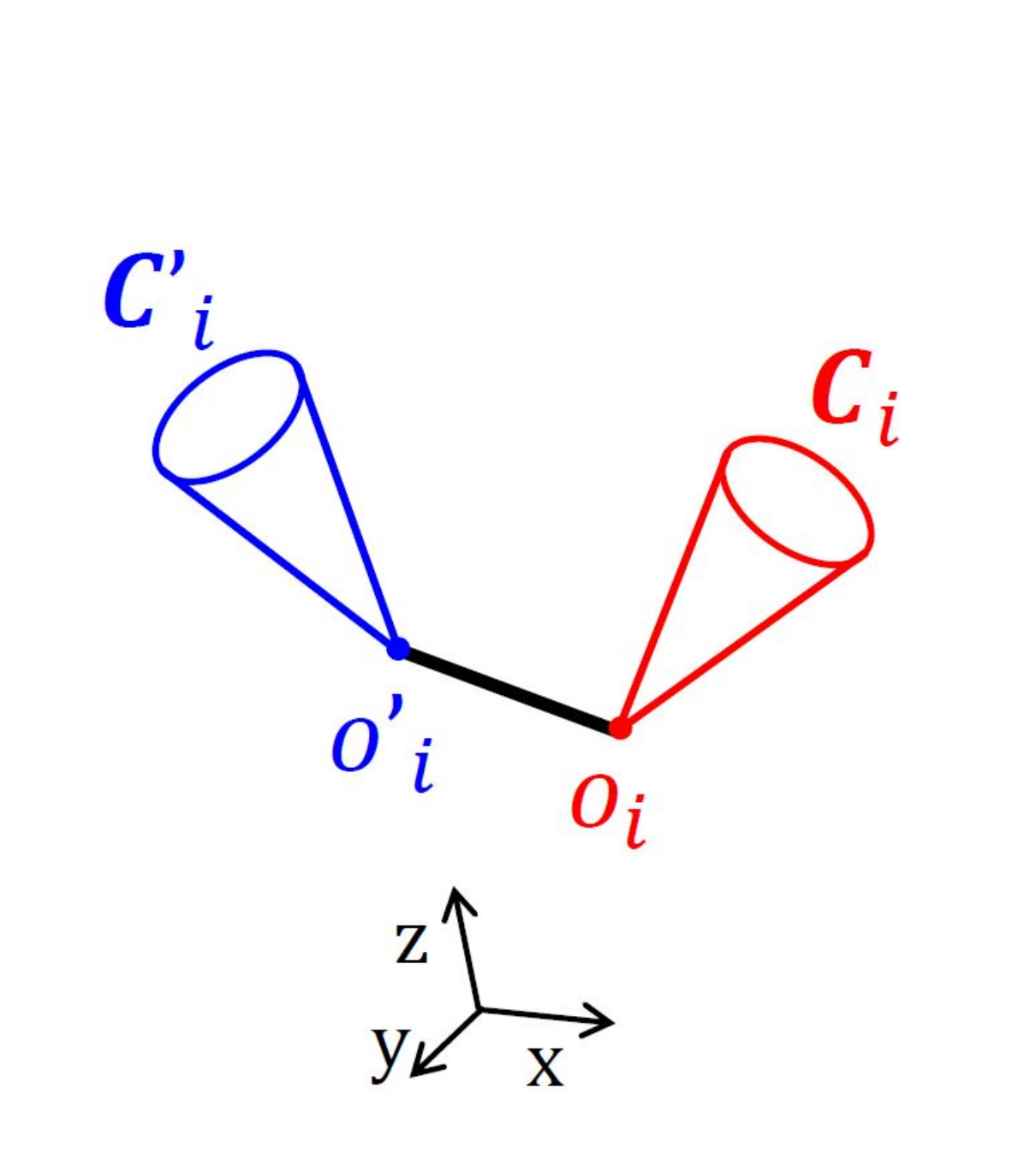}
			\caption{virtual 2-perspective-camera rig for $i$-th match.}
		\end{subfigure}
		\caption{Application of the six-point method to the generalized camera model.}
		\label{fig:Application_six-point}
	\end{figure}
	
A point match of a spatial point captured by a generalized camera can be described by two rays. As shown in \cref{fig:Application_six-point}(a), the intersection of two rays demonstrates that the same spatial point has been observed by a generalized camera in two views. For the $i$-th point match, denote $o_i$ and $o'_i$ as points lying on these two rays. Then, we can define a virtual 2-perspective-camera rig whose reference is consistent with the original generalized camera, as shown in \cref{fig:Application_six-point}(b). The optical centers of the perspective cameras are $o_i$ and $o'_i$, expressed in the generalized camera reference, respectively. The orientations of the virtual cameras are somewhat arbitrary since the virtual cameras can have a large FOV to capture the spatial point for the $i$-th match. Note that the extrinsic parameters of the virtual cameras are not coupled with the unknown relative pose of the generalized camera.
	
Based on the previous analysis, we have the following conclusions: (1) Each point match of a generalized camera can be equivalently captured by two virtual perspective cameras. (2) The construction (and thus the extrinsic parameters) of the virtual cameras is not unique. For possible virtual cameras, their observations for the point match are determined by corresponding extrinsic parameters. As a result, the extrinsic parameters of virtual cameras will not change the relative pose estimation results. (3) Since we focus on six-point matches, a virtual 12-perspective-camera rig should be constructed. Then the proposed generic six-point solver can be applied.

\section{Configuration Enumeration for Six-Point Solvers}
In Section~3.4 of the paper, two practical configurations called \texttt{inter-camera} and \texttt{intra-camera} are investigated for the six-point problem for two-camera rigs. Please refer to Fig.~3 in the paper. Then a question naturally appears as follows:
\begin{itemize}
	\item How to enumerate all configurations of six-point problems for multi-camera systems?
\end{itemize}

\subsection{Problem Definition}
\label{sec:problem_definition}
\begin{figure}[tbp]
	\centering
	\begin{subfigure}{0.41\linewidth}
		\includegraphics[width=1.0\linewidth]{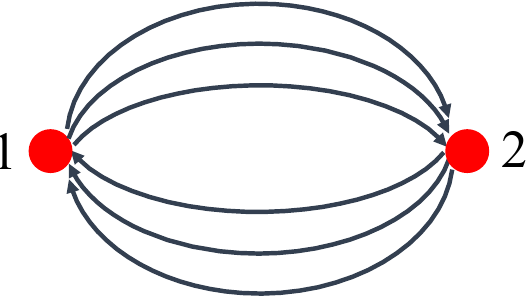}
		\caption{Inter-camera case.}
	\end{subfigure}
	\hfill \
	\begin{subfigure}{0.54\linewidth}
		\includegraphics[width=1.0\linewidth]{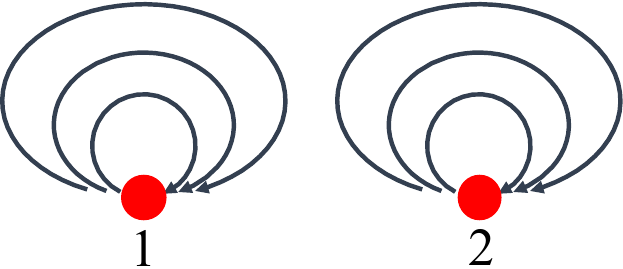}
		\caption{Intra-camera case.}
	\end{subfigure}
	\caption{Directed graph representation of the \texttt{inter-camera} and \texttt{intra-camera} cases.}
	\label{fig:graph_representation}
\end{figure}

A configuration of point correspondences (PCs) can be modeled by a directed graph (also called a digraph) by the following procedure. 
The vertex $i$ in the graph represents the single camera $i$ in a multi-camera system. An edge from vertex $i$ to $j$ in the graph represents a PC from camera $i$ in view 1 to camera $j$ in view 2. For example, the \texttt{inter-camera} and \texttt{intra-camera} cases in Fig.~3 in the paper can be represented by \cref{fig:graph_representation}. Then enumerating all the configurations can be equivalently converted to a graphical enumeration problem~\cite{harary1973graphical}.

{\bf Problem Reformulation}: We aim to count the $n$-edge directed graphs. The graphs do not have an isolated vertex. The graphs allow for the presence of loops, which denote edges connecting a vertex to itself, and multiple edges, indicating the existence of more than one edge connecting the same pair of vertices.

How to count and enumerate distinct graphs for a given edge number $n$? Two graphs are distinct if and only if two conditions are met: (1) they are not isomorphic, (2) they cannot be converted to each other by reversing all edge directions of one graph. A graph $\mathcal{G}$ comprises a pair $(\mathcal{V}, \mathcal{E})$, where $\mathcal{V}$ represents the set of vertices, and $\mathcal{E}$ represents the set of edges. In this paper, we are particularly interested in $n = 6$, \emph{i.e.}, a minimal configuration of six PCs.

{\bf Isomorphism} [Section~5 of~\cite{guichard2023combinatorics}]: Two graphs $\mathcal{G}_1 = (\mathcal{V}_1, \mathcal{E}_1)$ and $\mathcal{G}_2 = (\mathcal{V}_2, \mathcal{E}_2)$ are isomorphic if there exists a bijection $f: \mathcal{V}_1 \to \mathcal{V}_2$ such that a directed edge $(u,v) \in \mathcal{E}_1$ if and only if $(f(u), f(v)) \in \mathcal{E}_2$. Furthermore, the repetition numbers of $(u, v)$ and $(f(u), f(v))$ remain the same if multiple edges or loops are allowed. This bijection $f$ is termed an isomorphism.

{\bf Edge-Reversion Symmetry}: This is a non-standard symmetry which is meaningful in our problem. A solver is applicable to a relative pose estimation problem whether it can determine the relative pose from view~1 to view~2 or from view~2 and view~1.

In the reformulated problem, the graph isomorphism is used to exclude the symmetry of camera indices. In other words, swapping camera indices does not change the configuration. The edge-reversion symmetry is used to exclude the symmetry of two views.
A directed graph $\mathcal{G}$ with edges $\mathcal{E} = \{(u_i, v_i)\}_{i=1}^n$ and no isolated vertex can be represented by
\begin{align}
	E = 
	\left[
	\begin{array}{cccc}
		u_1 & u_2 & \cdots & u_n\\v_1 & v_2 & \cdots & v_n
	\end{array}
	\right]_{2\times n}
\end{align}
It is obvious that rearranging columns does not change the graph at all. 
In addition, a permutation of vertices in $E$ causes an isomorphic graph. Swapping two rows corresponds to reversing the edge directions of this graph. In this paper, we write $E_1 \cong E_2$ if $E_2$ can be obtained by rearrangement of columns, swapping rows, and swapping vertices of $E_1$.

{\bf Explanation of Isomorphism and Reversion Symmetry}: When the first row is $\{1, 2, 3\}$ and the second row contain $\{1, 2, 3\}$, there are $6$ permutations, including
\begin{align*}
	& E_1 = \left[
	\begin{array}{ccc}
		1 & 2 & 3 \\1 & 2 & 3
	\end{array}
	\right], \ \ 
	E_2 = \left[
	\begin{array}{ccc}
		1 & 2 & 3 \\1 & 3 & 2
	\end{array}
	\right], \\
	& E_3 = \left[
	\begin{array}{ccc}
		1 & 2 & 3 \\2 & 1 & 3
	\end{array}
	\right], \ \
	E_4 = \left[
	\begin{array}{ccc}
		1 & 2 & 3 \\2 & 3 & 1
	\end{array}
	\right], \\
	& E_5 = \left[
	\begin{array}{ccc}
		1 & 2 & 3 \\3 & 1 & 2
	\end{array}
	\right], \ \
	E_6 = \left[
	\begin{array}{ccc}
		1 & 2 & 3 \\3 & 2 & 1
	\end{array}
	\right].
\end{align*}
It can be verified that $E_3$ can be converted to $E_2$ by swapping vertices $1$ and $3$; $E_6$ can be converted to $E_2$ by swapping vertices $1$ and $2$. $E_5$ can be converted to $E_4$ by swapping its two rows. As a result, there are $3$ distinct graphs. The distinct graphs with isomorphism and edge-reversion symmetry are $E_1$, $E_2 \cong E_3 \cong E_6$, and $E_4 \cong E_5$.

\subsection{A Combinatorics Solution to Enumeration}
The problem is related to combinatorics and group theory. The P{\'o}lya enumeration theorem can be applied to solve this problem~\cite{guichard2023combinatorics,OEIS-A364088}. To apply this theorem, we need to generate the cycle index of the permutation group $S_n$ acting on $n$ elements.  Then we obtain the number of isomorphic graphs from this cycle index of $S_n$. 

In addition to theoretical analysis, we can enumerate all the distinct graphs in a recursive way by computer programming. We use about $50$ lines of Matlab codes to determine the distinct graphs, as shown in \cref{lst:graph}.

\subsection{Statistics of Cases and Solvers}
The number $D_n$ of distinct graphs for different edge numbers $n$ are shown in \cref{tab:number_of_graphs}. When $n = 6$ and $7$, the number of cases for different camera numbers are shown in \cref{tab:number_of_cases_6} and \cref{tab:number_of_cases_7}, respectively.

Since this paper focuses on six-point methods, we are particularly interested in the situation of $n = 6$. When $n = 6$, we classify its $5953$ cases into $6$~match types as below.

{
\renewcommand{\thelstlisting}{\arabic{lstlisting}}
\scriptsize
\begin{lstlisting}[caption=Enumerate Distinct Graphs Using Matlab, label={lst:graph}] 
function G_curr = generate_distict_graph(edge_num)
G_curr = {};
for k = 1:edge_num
  G_prev = G_curr;
  G_curr = generate_graph_recursion(G_prev, k);
end
		
function G_curr = generate_graph_recursion(G_prev, k)
if k == 1
  G_curr = cell(1, 2);
  G_curr{1} = struct('E', [1; 1]); 
  G_curr{2} = struct('E', [1; 2]); 
  return;
end
curr_sz = 0;
G_curr = {};
for i = 1:numel(G_prev)
  E0 = G_prev{i}.E;
  edge_all = generate_edge_set(E0);
  for j = 1:size(edge_all, 2)
    E = [E0, edge_all(:, j)];
    G = digraph(E(1, :), E(2, :));
    H = digraph(E(2, :), E(1, :));
    if ~is_in_set(G, H, G_curr, curr_sz)
     curr_sz = curr_sz + 1;
     G_curr{curr_sz} = struct('E', E, 'G', G);
    end
  end
end
fprintf('#edge: %d, #found-graphs: %d\n', k, curr_sz);
		
function edge_all = generate_edge_set(E)
s0 = unique(E(:));
s = [s0(:); max(s0)+1];
n = numel(s);
edge_all = zeros(2, n*n+1);
i1 = kron(1:n, ones(1,n));
i2 = kron(ones(1,n), 1:n);
edge_all(:, 1:n^2) = [s(i1), s(i2)]'; 
edge_all(:, end) = max(s0)+[1; 2];
		
function flag = is_in_set(G, H, G_set, curr_sz)
flag = false;
  for ii = 1:curr_sz
    G1 = G_set{ii}.G;
    if (isisomorphic(G1, G) || isisomorphic(G1, H))
      flag = true;
    return;
  end
end
\end{lstlisting}
}

\begin{table*}[bp]
	\caption{The number $D_n$ of distinct graphs given $n$ edges.}
	\begin{center}
		\setlength{\tabcolsep}{3.9pt}{
			\begin{tabular}{lcccccccccc}
				\toprule
				\#edge $n$ & 1 & 2 & 3 & 4 & 5 & 6 & 7 & 8 & 9 & 10\\
				\midrule
				\#distinct graph & 2 & 9 & 37 & 186 & 985 & 5953 & 38689 & 271492 & 2016845 & 15767277 \\
				\bottomrule
			\end{tabular}
		}
	\end{center}
	\label{tab:number_of_graphs}
\end{table*}

\begin{table*}[tbp]
	\caption{The number of cases for different camera numbers given $6$ edges. Totally there are $5953$ cases.}
	\begin{center}
		\setlength{\tabcolsep}{5.9pt}{
			\begin{tabular}{lcccccccccccc}
				\toprule
				\#camera & 1 & 2 & 3 & 4 & 5 & 6 & 7 & 8 & 9 & 10 & 11 & 12\\
				\midrule
				\#cases & 1 & 29 & 270 & 1029 & 1776 & 1630 & 853 & 280 & 66 & 15 & 3 & 1 \\
				\bottomrule
			\end{tabular}
		}
	\end{center}
	\label{tab:number_of_cases_6}
\end{table*}

\begin{table*}[t]
	\caption{The number of cases for different camera numbers given $7$ edges. Totally there are $38689$ cases.}
	\begin{center}
		\setlength{\tabcolsep}{3.65pt}{
			\begin{tabular}{lcccccccccccccc}
				\toprule
				\#camera & 1 & 2 & 3 & 4 & 5 & 6 & 7 & 8 & 9 & 10 & 11 & 12 & 13 & 14\\
				\midrule
				\#cases & 1 & 39 & 568 & 3316 & 8599 & 11516 & 8787 & 4170 & 1296 & 312 & 66 & 15 & 3 & 1 \\
				\bottomrule
			\end{tabular}
		}
	\end{center}
	\label{tab:number_of_cases_7}
\end{table*}

\begin{table}[t]
	\caption{The number of solutions and the number of solvers for different match types when $n = 6$. Totally there are $5953$ cases.}
	\begin{center}
		\setlength{\tabcolsep}{6pt}{
			\begin{tabular}{lcccccc}
				\toprule
				match type & $6\cup \phi$ & $5\cup x$ & $4\cup x$ & $3\cup 3$ & $3\cup x$ & $\cup_i x_i$ \\
				\midrule
				\#case & 2 & 9 & 63 & 7 & 412 & 5460 \\
				\midrule
				\#equ. in $\mathcal{E}_1$ & 0 & 10 & 14 & 15 & 15 & 15 \\
				\#equ. in $\mathcal{E}_2$ & 20 & 10 & 4 & 2 & 1 & 0  \\
				\midrule
				\#solution & 0 & 20 & 40 & 48 & 56 & 64 \\
				\bottomrule
			\end{tabular}
		}
	\end{center}
	\label{tab:solution_number}
\end{table}

\begin{itemize}
	\item Match type $6\cup \phi$: All the 6 edges are same. There are $2$ cases belonging to this match type, including
	\begin{align*}
		& E_1 = \left[
		\begin{array}{cccccc}
			1 \ \ & 1 \ \ & 1 \ \ & 1 \ \ & 1 \ \ & 1 \\
			1 \ \ & 1 \ \ & 1 \ \ & 1 \ \ & 1 \ \ & 1
		\end{array}
		\right], \\
		& E_2 = \left[
		\begin{array}{cccccc}
			1 \ \ & 1 \ \ & 1 \ \ & 1 \ \ & 1 \ \ & 1 \\
			2 \ \ & 2 \ \ & 2 \ \ & 2 \ \ & 2 \ \ & 2
		\end{array}
		\right].
	\end{align*}
	These two cases essentially correspond to monocular cameras. Since $6$ PCs provide an over-determined equation system, the equation system has no solution.
	\item Match type $5\cup x$: The maximal repetitive number of multiple edges is $5$. For example,
	\begin{align*}
		& E_3 = \left[
		\begin{array}{cccccc}
			1 \ \ & 1 \ \ & 1 \ \ & 1 \ \ & 1 \ \ & 2 \\
			1 \ \ & 1 \ \ & 1 \ \ & 1 \ \ & 1 \ \ & 3
		\end{array}
		\right].
	\end{align*}
	\item Match type $4\cup x$: The maximal repetitive number of multiple edges is $4$. For example,
	\begin{align*}
		& E_4 = \left[
		\begin{array}{cccccc}
			1 \ \ & 1 \ \ & 1 \ \ & 1 \ \ & 2 \ \ & 4 \\
			1 \ \ & 1 \ \ & 1 \ \ & 1 \ \ & 3 \ \ & 5
		\end{array}
		\right].
	\end{align*}
	\item Match type $3\cup 3$: The maximal repetitive number of multiple edges is $3$, and there are two groups of multiple edges satisfying this condition. For example, the previously defined \texttt{inter-camera} and \texttt{intra-camera} cases belong to this match type. They are
	\begin{align*}
		& E_5 = \left[
		\begin{array}{cccccc}
			1 \ \ & 1 \ \ & 1 \ \ & 2 \ \ & 2 \ \ & 2 \\
			2 \ \ & 2 \ \ & 2 \ \ & 1 \ \ & 1 \ \ & 1
		\end{array}
		\right], \\
		& E_6 = \left[
		\begin{array}{cccccc}
			1 \ \ & 1 \ \ & 1 \ \ & 2 \ \ & 2 \ \ & 2 \\
			1 \ \ & 1 \ \ & 1 \ \ & 2 \ \ & 2 \ \ & 2
		\end{array}
		\right].
	\end{align*}
	\item Match type $3\cup x$: The maximal repetitive number of multiple edges is $3$, and there is only one group of multiple edges satisfying this condition. For example,
	\begin{align*}
		& E_7 = \left[
		\begin{array}{cccccc}
			1 \ \ & 1 \ \ & 1 \ \ & 2 \ \ & 4 \ \ & 6 \\
			1 \ \ & 1 \ \ & 1 \ \ & 3 \ \ & 5 \ \ & 7
		\end{array}
		\right].
	\end{align*}
	\item Match type $\cup_i x_i$: The maximal repetitive number of multiple edges is less than or equal to $2$. For example,
	\begin{align*}
		& E_8 = \left[
		\begin{array}{cccccc}	
			1 \ \ & 3 \ \ & 5 \ \ & 7 \ \ & 9 \ \ & 11 \\
			2 \ \ & 4 \ \ & 6 \ \ & 8 \ \ & 10 \ \ & 12
		\end{array}
		\right].
	\end{align*}
	Most of the cases belong to this match type. In the previous section, we call it the generic case.
\end{itemize}

For different match types, the numbers of equations in $\mathcal{E}_1$ and $\mathcal{E}_2$ are different. We found that cases belonging to the same match type have preciously the same number of solutions. \cref{tab:solution_number} shows the number of solutions for different match types in six-point problems for multi-camera systems.

\section{\label{sec:DegenerateConfigurations}Degenerate Configurations} 
For two-camera rigs, there are degenerate cases for \texttt{6pt+cayl/quat+generic}. The devised solvers \texttt{6pt+cayl/quat+inter} and \texttt{6pt+cayl/quat+intra} have better performance than the generic solver in those cases. We also introduce two degenerate cases for these two solvers for two-camera rigs: (1) For inter-camera case, a two-camera rig undergoes pure translation while the baselines of the two cameras are parallel to the translation direction, (2) For intra-camera case, a two-camera rig undergoes pure translation or the cameras move along concentric circles. In these critical configurations, the metric scale of the translation is unobtainable.

\section{Experiments}
\label{sec:experiment}
In this section, we add more experiments about the 6DOF relative pose estimation of a multi-camera system. As described in the paper, all the solvers are implemented in C++. The source codes for \texttt{17pt-Li}~\cite{li2008linear} and \texttt{8pt-Kneip}~\cite{kneip2014efficient} are adopted from the OpenGV library~\cite{kneip2014opengv}. The source codes for \texttt{6pt-Stew{\'e}nius}~\cite{stewenius2005solutions} is adopted from the PoseLib library~\cite{PoseLib}. The efficiency and numerical stability of the proposed solvers are shown in \cref{sec:numericalstability}. Expected runtime vs. inlier ratio accounting for numerical stability is demonstrated in \cref{sec:expectedRuntime}. 
More experiments on synthetic data are shown in \cref{sec:SyntheticExperiments}. The real-world experiments on \texttt{KITTI} and \texttt{nuScenes} datasets are shown in \cref{sec:KITTIexperiments} and \cref{sec:nuScenesexperiments}, respectively. 

\subsection{\label{sec:numericalstability}Efficiency and Numerical Stability}
The runtimes of the solvers for multi-camera systems are assessed using an Intel(R) Core(TM) i7-7800X 3.50~GHz processor. The average runtime of the solvers over $10,000$ runs is shown in \cref{tab:SolverTime_generalized}. As a linear solver, \texttt{17pt-Li} is the most efficient one. All the proposed solvers have runtimes of $0.8 \sim 1.8$ milliseconds. The solvers \texttt{6pt-Our-inter56}, \texttt{6pt-Our-inter48}, and \texttt{6pt-Our-intra} are more efficient than the seminal \texttt{6pt-Stew{\'e}nius} among the minimal solvers.
\begin{table}[tbp]
	\caption{Comparison of solver runtime for multi-camera systems (unit:~$\mu s$).}
	\begin{center}
		\resizebox{\linewidth}{!}{
			\begin{tabular}{lccccccc}
				\toprule
				Methods & {17pt-Li}~\cite{li2008linear} & {8pt-Kneip}~\cite{kneip2014efficient} &  {6pt-Stew{\'e}nius}~\cite{stewenius2005solutions} & {6pt-Our-generic} & {6pt-Our-inter56} & {6pt-Our-inter48} & {6pt-Our-intra} \\
				\midrule
				Runtime & 43.3 & 102.0 & 1214.9 & 1724.6 & 1053.4 & 827.8 &  774.1 \\
				\bottomrule
		\end{tabular}}
	\end{center}
	\label{tab:SolverTime_generalized}
\end{table}

\begin{figure*}[tbp]
	\centering
	\includegraphics[width=0.94\linewidth]{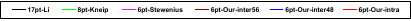}\\
	\centering
	\begin{subfigure}{0.32\linewidth}
		\includegraphics[width=1.0\linewidth]{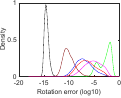}
		\caption{Revolution error $\varepsilon_{R}$.}
	\end{subfigure}
	\hfill
	\begin{subfigure}{0.32\linewidth}
		\includegraphics[width=1.0\linewidth]{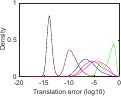}
		\caption{Translation error $\varepsilon_{\mathbf{t}}$.}
	\end{subfigure}
	\hfill
	\begin{subfigure}{0.32\linewidth}
		\includegraphics[width=1.0\linewidth]{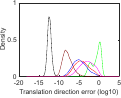}
		\caption{Translation dir. error $\varepsilon_{\mathbf{t},\text{dir}}$.}
	\end{subfigure}
	\caption{Probability density functions of the relative pose estimation error on noise-free observations under the first scenario. We designed a simulated two-camera rig comprising two perspective cameras. The orientations of two perspective cameras are roughly forward-facing with random perturbation. The horizontal axis corresponds to the $\log_{10}$ estimated errors, and the vertical axis is the empirical probability density.}
 \label{fig:NumericalStability_2cameras_kitti_our6pt}
\end{figure*}

\begin{figure*}
	\centering
	\includegraphics[width=0.60\linewidth]{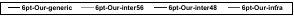}\\
	\centering
	\begin{subfigure}{0.32\linewidth}
		\includegraphics[width=1.0\linewidth]{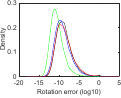}
		\caption{Rotation error $\varepsilon_{\R}$.}
	\end{subfigure}
	\hfill
	\begin{subfigure}{0.32\linewidth}
		\includegraphics[width=1.0\linewidth]{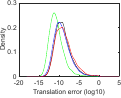}
		\caption{Translation error $\varepsilon_{\mathbf{t}}$.}
	\end{subfigure}
	\hfill
	\begin{subfigure}{0.32\linewidth}
		\includegraphics[width=1.0\linewidth]{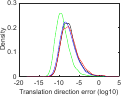}
		\caption{Translation dir. error $\varepsilon_{\mathbf{t},\text{dir}}$.}
	\end{subfigure}
	\centering
	\includegraphics[width=0.45\linewidth]{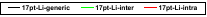}\\
	\centering
	\begin{subfigure}{0.3\linewidth}
		\includegraphics[width=1.0\linewidth]{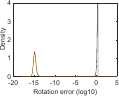}
		\caption{Rotation error $\varepsilon_{\R}$.}
	\end{subfigure}
	\hfill
	\begin{subfigure}{0.32\linewidth}
		\includegraphics[width=1.0\linewidth]{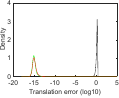}
		\caption{Translation error $\varepsilon_{\mathbf{t}}$.}
	\end{subfigure}
	\hfill
	\begin{subfigure}{0.32\linewidth}
		\includegraphics[width=1.0\linewidth]{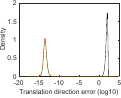}
		\caption{Translation dir. error $\varepsilon_{\mathbf{t},\text{dir}}$.}
	\end{subfigure}
	\centering
	\includegraphics[width=0.5\linewidth]{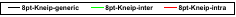}\\
	\centering
	\begin{subfigure}{0.32\linewidth}
		\includegraphics[width=1.0\linewidth]{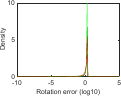}
		\caption{Rotation error $\varepsilon_{\R}$.}
	\end{subfigure}
	\hfill
	\begin{subfigure}{0.32\linewidth}
		\includegraphics[width=1.0\linewidth]{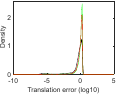}
		\caption{Translation error $\varepsilon_{\mathbf{t}}$.}
	\end{subfigure}
	\hfill
	\begin{subfigure}{0.32\linewidth}
		\includegraphics[width=1.0\linewidth]{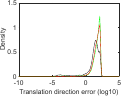}
		\caption{Translation dir. error $\varepsilon_{\mathbf{t},\text{dir}}$.}
	\end{subfigure}
	\centering
	\includegraphics[width=0.55\linewidth]{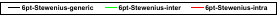}\\
	\centering
	\begin{subfigure}{0.32\linewidth}
		\includegraphics[width=1.0\linewidth]{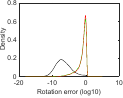}
		\caption{Rotation error $\varepsilon_{\R}$.}
	\end{subfigure}
	\hfill
	\begin{subfigure}{0.32\linewidth}
		\includegraphics[width=1.0\linewidth]{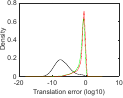}
		\caption{Translation error $\varepsilon_{\mathbf{t}}$.}
	\end{subfigure}
	\hfill
	\begin{subfigure}{0.32\linewidth}
		\includegraphics[width=1.0\linewidth]{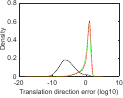}
		\caption{Translation dir. error $\varepsilon_{\mathbf{t},\text{dir}}$.}
	\end{subfigure}
	\caption{Probability density functions of the relative pose estimation error on noise-free observations under the second scenario. We designed a simulated multi-camera system comprising twelve omnidirectional cameras. The extrinsic parameters, including orientation and position, are totally random. The horizontal axis corresponds to the $\log_{10}$ estimated errors, and the vertical axis is the empirical probability density.}
	\label{fig:NumericalStability_12cameras_our6pt}
\end{figure*}

To test the numerical stability, we design two simulated scenarios. First, we design a simulated two-camera rig composed of two perspective cameras. The orientations of the two perspective cameras are roughly forward-facing with random perturbation. This setting is practical for autonomous driving with two front-facing cameras. Second, we design a simulated generalized camera composed of twelve omnidirectional cameras. The extrinsic parameters, including orientation and position, are totally random. 

\cref{fig:NumericalStability_2cameras_kitti_our6pt} 
illustrates the numerical stability of the solvers when applied to noise-free observations under the first scenario. The procedure is repeated for $10,000$ times on both inter-camera and intra-camera cases. The resulting empirical probability density functions are represented as a function of the $\log_{10}$ estimated errors. The numerical stability of a solver is determined by many factors, such as the problem's complexity, parameterization, equation system construction, solver generator, and implementation. Usually, more efficient solvers generated by effective methods have less round-off error and thus have better numerical stability. Among these solvers, the \texttt{17pt-Li} solver demonstrates the best numerical stability due to its status as a linear solver with fewer computations. Since the \texttt{8pt-Kneip} solver relies on iterative optimization, it is susceptible to becoming trapped in local minima. 
The \texttt{6pt-Stew{\'e}nius} solver does not work well with a simulated two-camera rig. This phenomenon is consistent with~\cite{kim2009motion}, which observes the \texttt{6pt-Stew{\'e}nius} solver does not work for most axial cameras. All the proposed solvers exhibit satisfactory numerical stability. Since the \texttt{6pt-Our-inter56} solver displays better numerical stability than the \texttt{6pt-Our-inter48} solver, we recommend \texttt{6pt-Our-inter56} as the default solver for inter-camera case in the follow-up experiments.

\cref{fig:NumericalStability_12cameras_our6pt} 
illustrates the numerical stability of the solvers when applied to noise-free observations under the second scenario. The numerical stability of the solvers is tested in generic-camera, inter-camera, and intra-camera configurations. The strategy of correspondence selection is described in \cref{sec:SynexperimentsGene}. For all three cases, the proposed solvers have satisfactory numerical stability. Among them, the \texttt{6pt-Our-inter56} solver has the best numerical stability. None of the comparison solvers can have satisfactory results for all the cases. Specifically, 
\begin{itemize}
	\item \texttt{17pt-Li} has satisfactory results for inter-camera and intra-camera cases. It does not work for generic case, because the essential matrix with scale ambiguity cannot be uniquely recovered due to the rank deficiency of the coefficient matrix. It can be proved by integer observations as that in~\cite{kim2009motion}. Thus, it is not surprising that the corresponding numerical stability is extraordinarily bad.
	\item \texttt{8pt-Kneip} does not have satisfactory results. The ground truth of rotation is random in the experiment settings, and this solver uses an identity matrix to initialize the rotation. As a result, its solutions determined by local optimization are prone to local minima.
	\item \texttt{6pt-Stew{\'e}nius} has satisfactory results for generic-camera configuration. Since this solver does not work for most axial cameras~\cite{kim2009motion}, it does not work well for inter-camera and intra-camera cases. 
\end{itemize}

\begin{figure}[tbp]
	\centering
	\begin{subfigure}{0.47\linewidth}
		\includegraphics[width=0.9\linewidth]{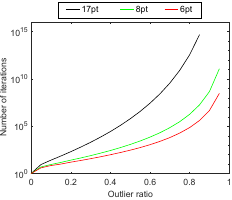}
		\caption{Iteration number without accounting for numerical stability.}
	\end{subfigure} 
	\begin{subfigure}{0.47\linewidth}
		\includegraphics[width=0.9\linewidth]{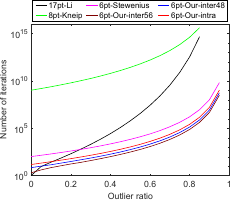}
		\caption{Iteration number accounting for numerical stability.}
	\end{subfigure}
	\caption{The relationship between the RANSAC iteration number and the outlier ratio in the context of achieving a success probability of $99\%$.}
	\label{fig:RANSAC_iteration_number}
\end{figure}

\subsection{\label{sec:expectedRuntime}Expected Runtime vs. Inlier Ratio Accounting for Numerical Stability}
The computational time of RANSAC is contingent upon both the runtime of the minimal solver and the number of iterations. For the RANSAC methods, the iteration number is determined by the probability of choosing at least one inlier set. This calculation implicitly assumes that an accurate solution can be obtained given an inlier set. However, this is not always true when considering the numerical stability of the minimal solvers.

Denote $p$ is the probability (usually set to 0.99) that at least one inlier set is sampled. $n$ is the minimum number of required data points in the minimal solver. For example, $s=6$ for the six-point solvers and $s=17$ for the 17-point solver. $\epsilon$ is the outlier ratio, \emph{i.e.}, the probability that any selected data point is an outlier. $N$ is the iteration number of RANSAC. In $N$ sampling times, the failure probability (one or more outliers are sampled in every sampling) is 
\begin{align}
	(1-(1-\epsilon)^s)^N = 1 - p,
\end{align}
and
\begin{align}
	N = \frac{\log (1-p)}{\log (1 - (1-\epsilon)^s)}.
\end{align}

When considering numerical stability, suppose $p_2$ is the probability of obtaining a sufficiently accurate solution given an inlier set. 
In $\hat{N}$ sampling times, the failure probability accounting for numerical stability is 
\begin{align}
	(1-(p_2(1-\epsilon))^s)^{\hat{N}} = 1 - p,
\end{align}
and
\begin{align}
	\hat{N} = \frac{\log (1-p)}{\log (1 - (p_2(1-\epsilon))^s)}.
\end{align}
Since $p_2 \in [0, 1]$ as a probability, we have $\hat{N} \ge N$ when accounting for numerical stability.

The left question is how to determine $p_2$ for a given minimal solver. For each minimal solver, we can empirically obtain the probability that the estimation errors $\varepsilon_{\R}$ and $\varepsilon_{\mathbf{t}}$ on noise-free observations are below specified thresholds. In this paper, we set the thresholds as $10^{-3}$. For the solvers \texttt{17pt-Li}~\cite{li2008linear}, \texttt{8pt-Kneip}~\cite{kneip2014efficient}, \texttt{6pt-Stew{\'e}nius}~\cite{stewenius2005solutions}, \texttt{6pt-Our-inter56}, \texttt{6pt-Our-inter48} and \texttt{6pt-Our-intra}, $p_2$ can be computed as 1.00, 0.09, 0.61, 0.99, 0.90 and 0.81, respectively. \cref{fig:RANSAC_iteration_number} shows RANSAC iteration numbers with respect to the outlier ratio for success probability $99\%$.

\subsection{\label{sec:SyntheticExperiments}Experiments on Synthetic Data}
Following the experiments about numerical stability, two simulated scenarios are designed and tested for the synthetic experiments, including scenarios for a two-camera rig and a generalized camera. The proposed solvers are evaluated against state-of-the-art solvers including  \texttt{17pt-Li}~\cite{li2008linear}, \texttt{8pt-Kneip}~\cite{kneip2014efficient}, and \texttt{6pt-Stew{\'e}nius}~\cite{stewenius2005solutions}.
\begin{figure*}[t]
	\centering
	\includegraphics[width=0.50\linewidth]{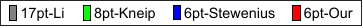}\\
	\centering
	\begin{subfigure}{0.32\linewidth}
		\includegraphics[width=1.0\linewidth]{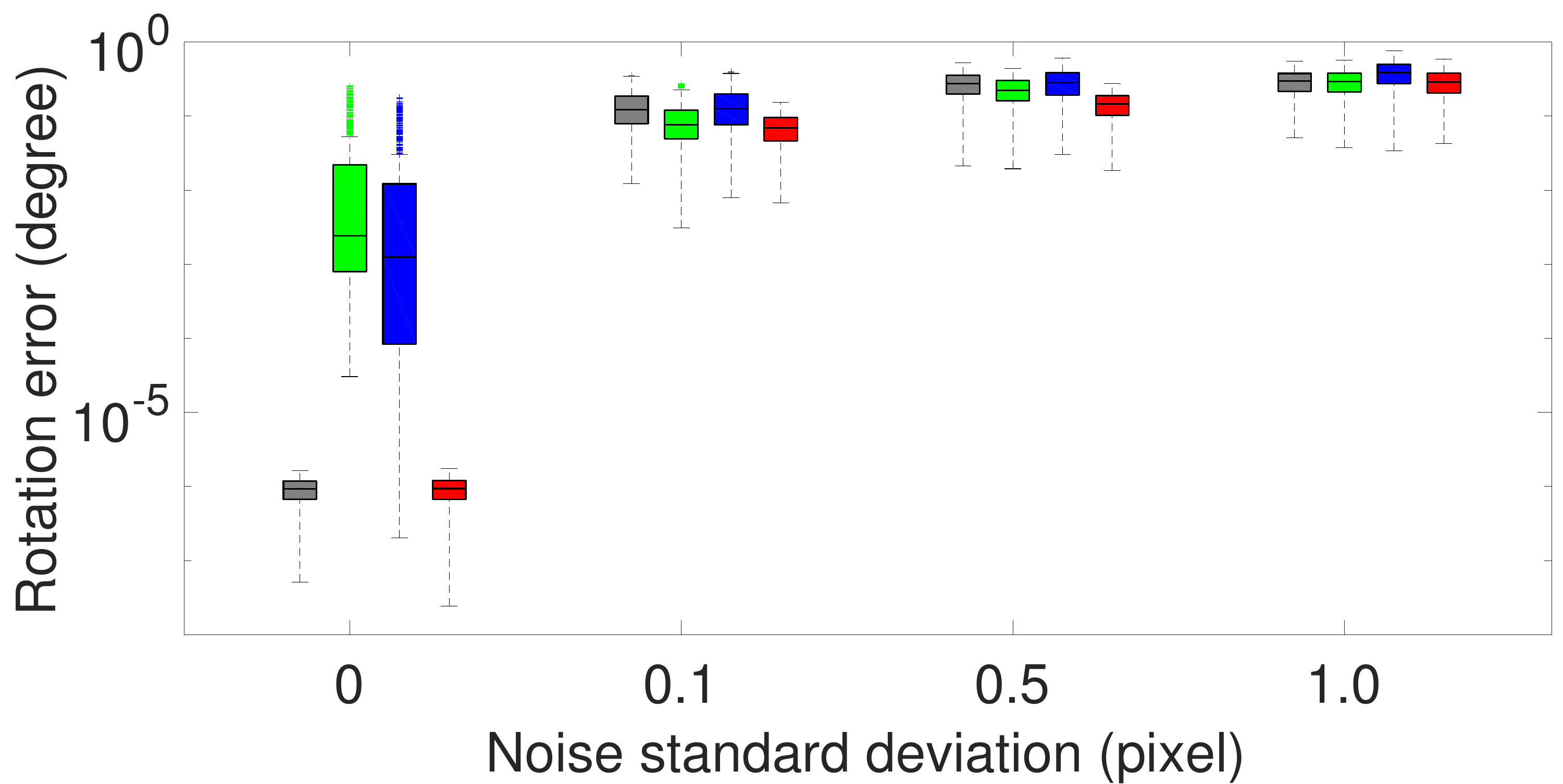}
		\caption{${\varepsilon_{\mathbf{R}}}$ for forward motion.}
	\end{subfigure}
	\hfill
	\begin{subfigure}{0.32\linewidth}
		\includegraphics[width=1.0\linewidth]{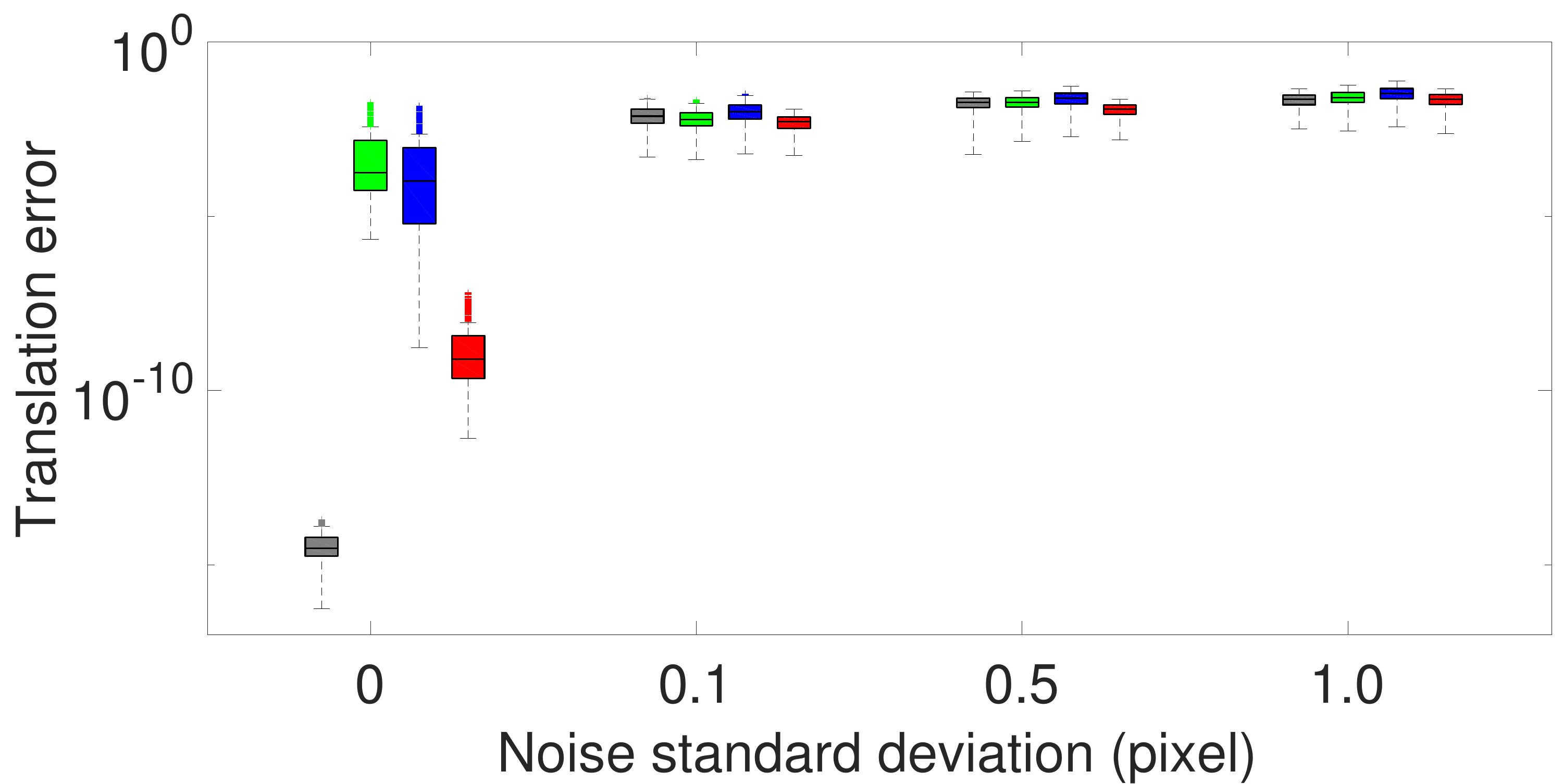}
		\caption{$\varepsilon_{\mathbf{t}}$ for forward motion.}
	\end{subfigure}
	\hfill
	\begin{subfigure}{0.32\linewidth}
		\includegraphics[width=1.0\linewidth]{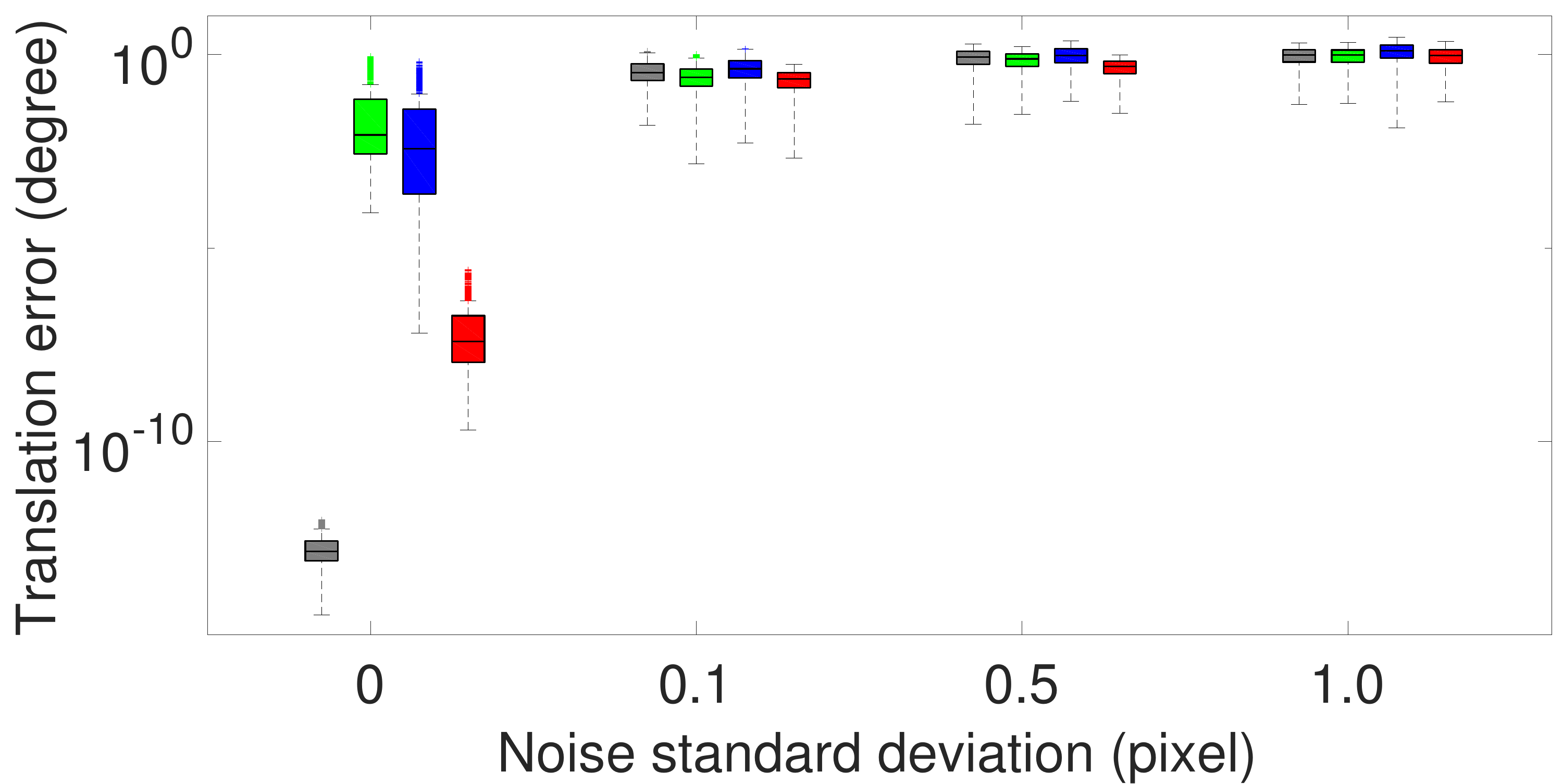}
		\caption{$\varepsilon_{\mathbf{t},\text{dir}}$ for forward motion.}
	\end{subfigure}
	\begin{subfigure}{0.32\linewidth}
		\includegraphics[width=1.0\linewidth]{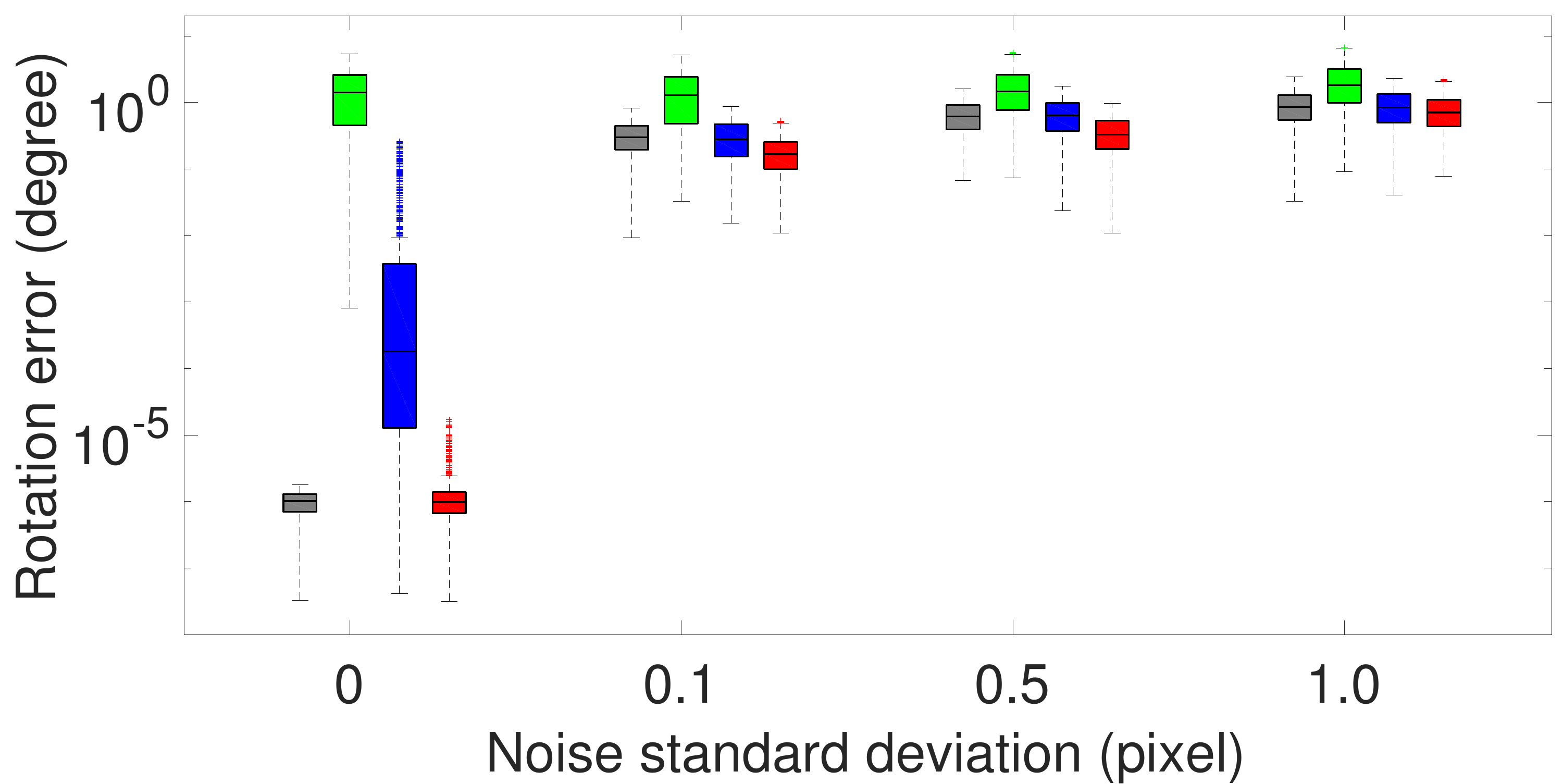}
		\caption{${\varepsilon_{\mathbf{R}}}$ for sideways motion.}
	\end{subfigure}
	\hfill
	\begin{subfigure}{0.32\linewidth}
		\includegraphics[width=1.0\linewidth]{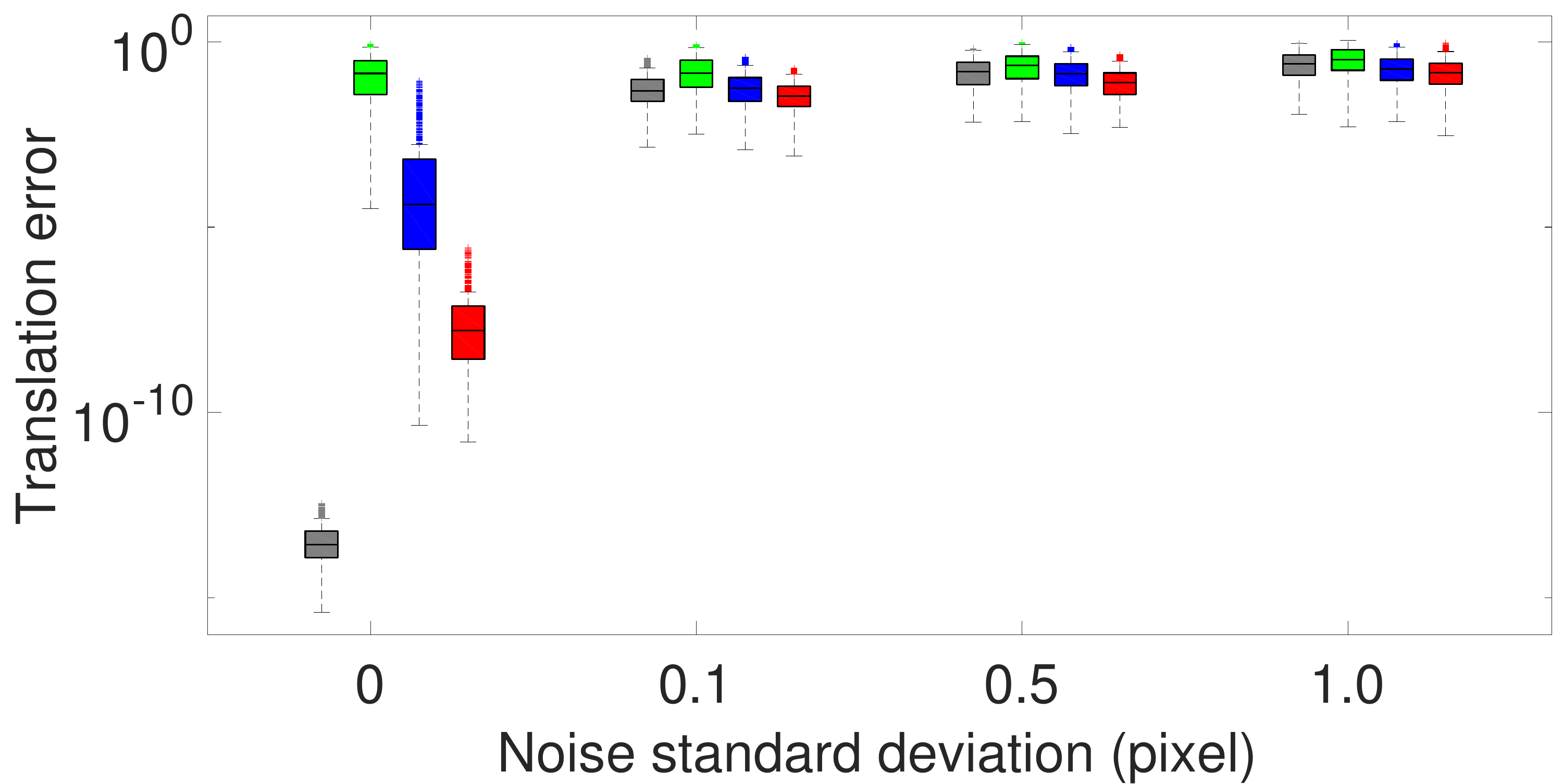}
		\caption{$\varepsilon_{\mathbf{t}}$ for sideways motion.}
	\end{subfigure}
	\hfill
	\begin{subfigure}{0.32\linewidth}
		\includegraphics[width=1.0\linewidth]{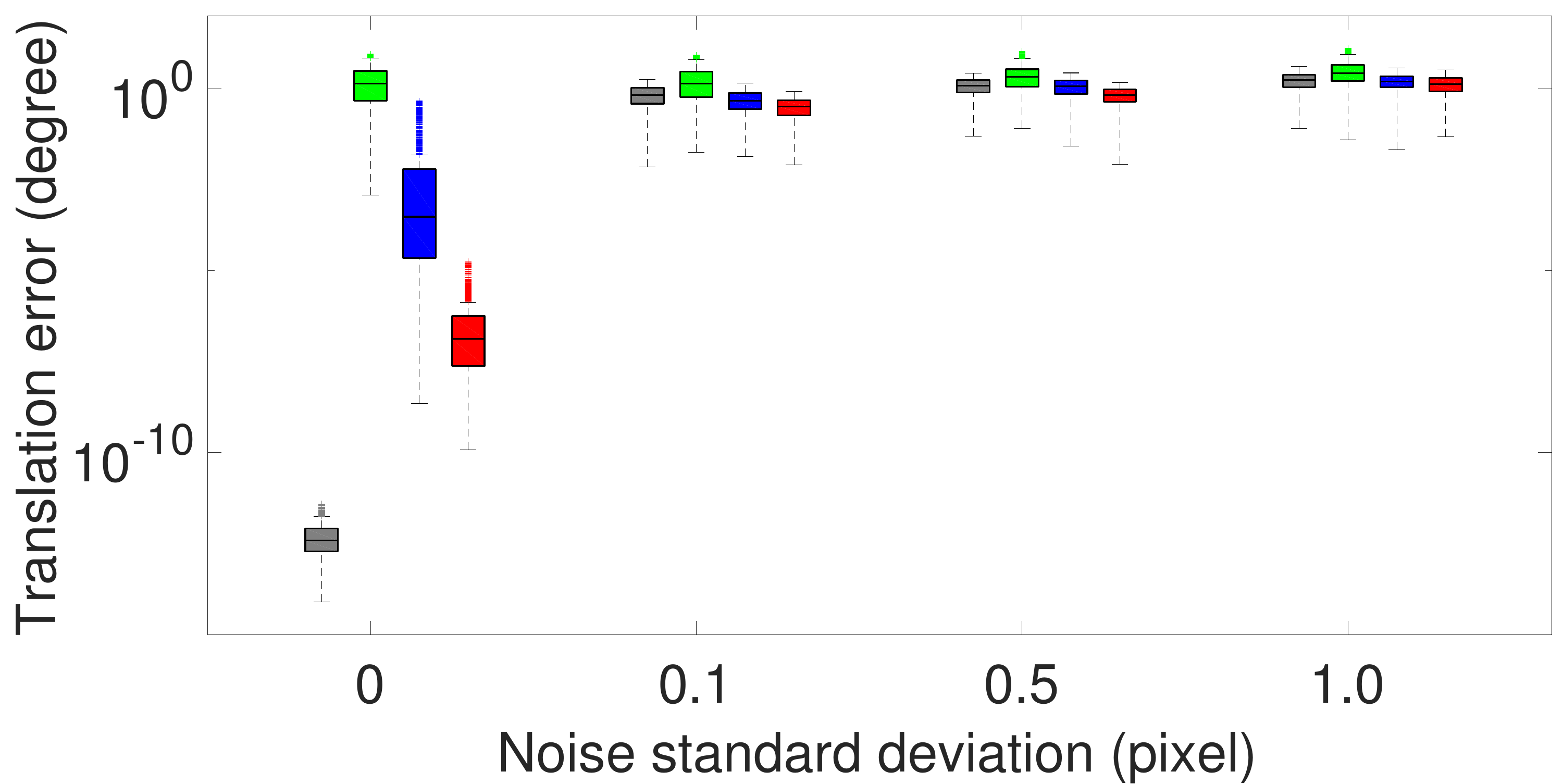}
		\caption{$\varepsilon_{\mathbf{t},\text{dir}}$ for sideways motion.}
	\end{subfigure}
	\begin{subfigure}{0.32\linewidth}
		\includegraphics[width=1.0\linewidth]{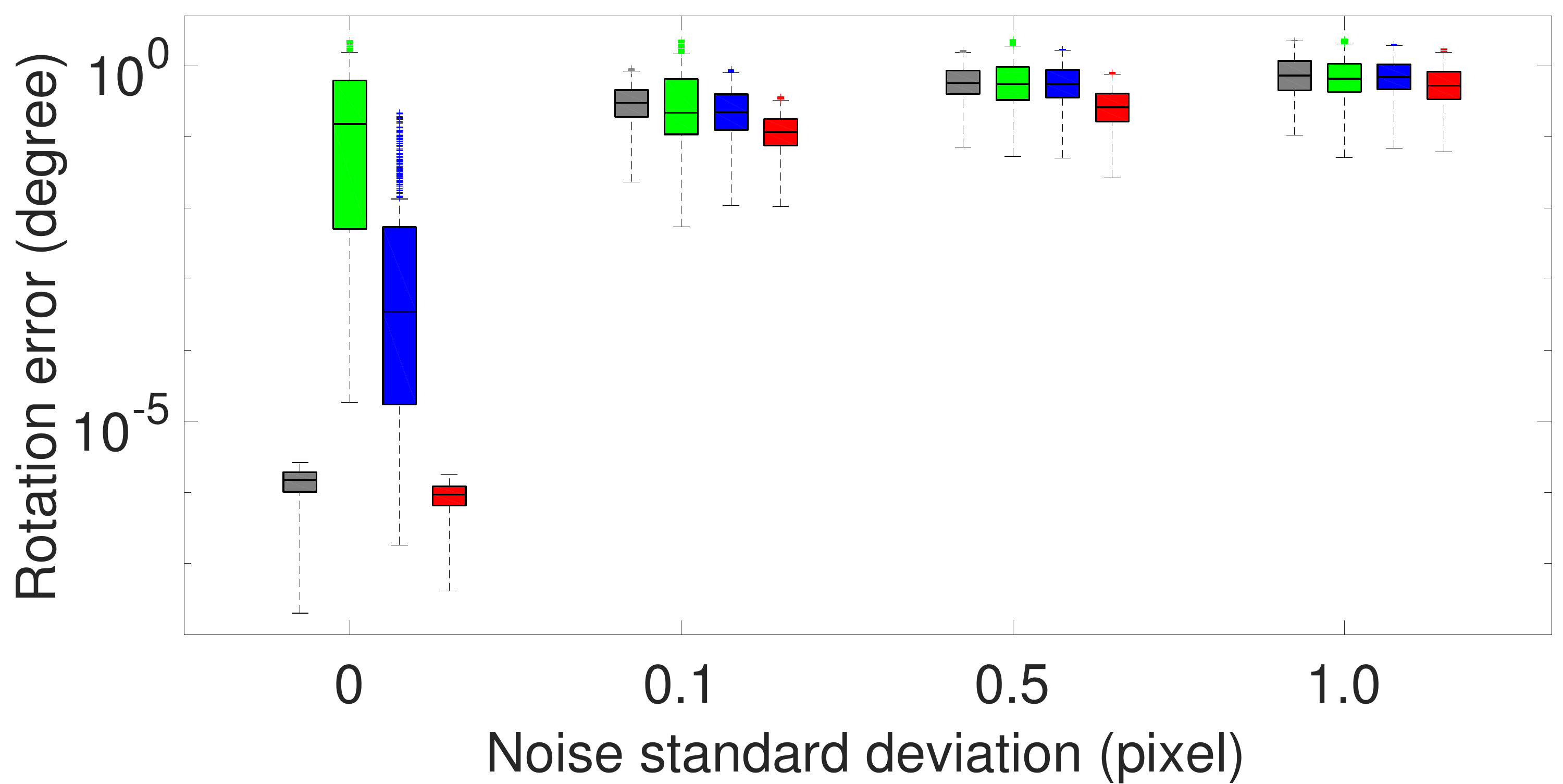}
		\caption{${\varepsilon_{\mathbf{R}}}$ for random motion.}
	\end{subfigure}
	\hfill
	\begin{subfigure}{0.32\linewidth}
		\includegraphics[width=1.0\linewidth]{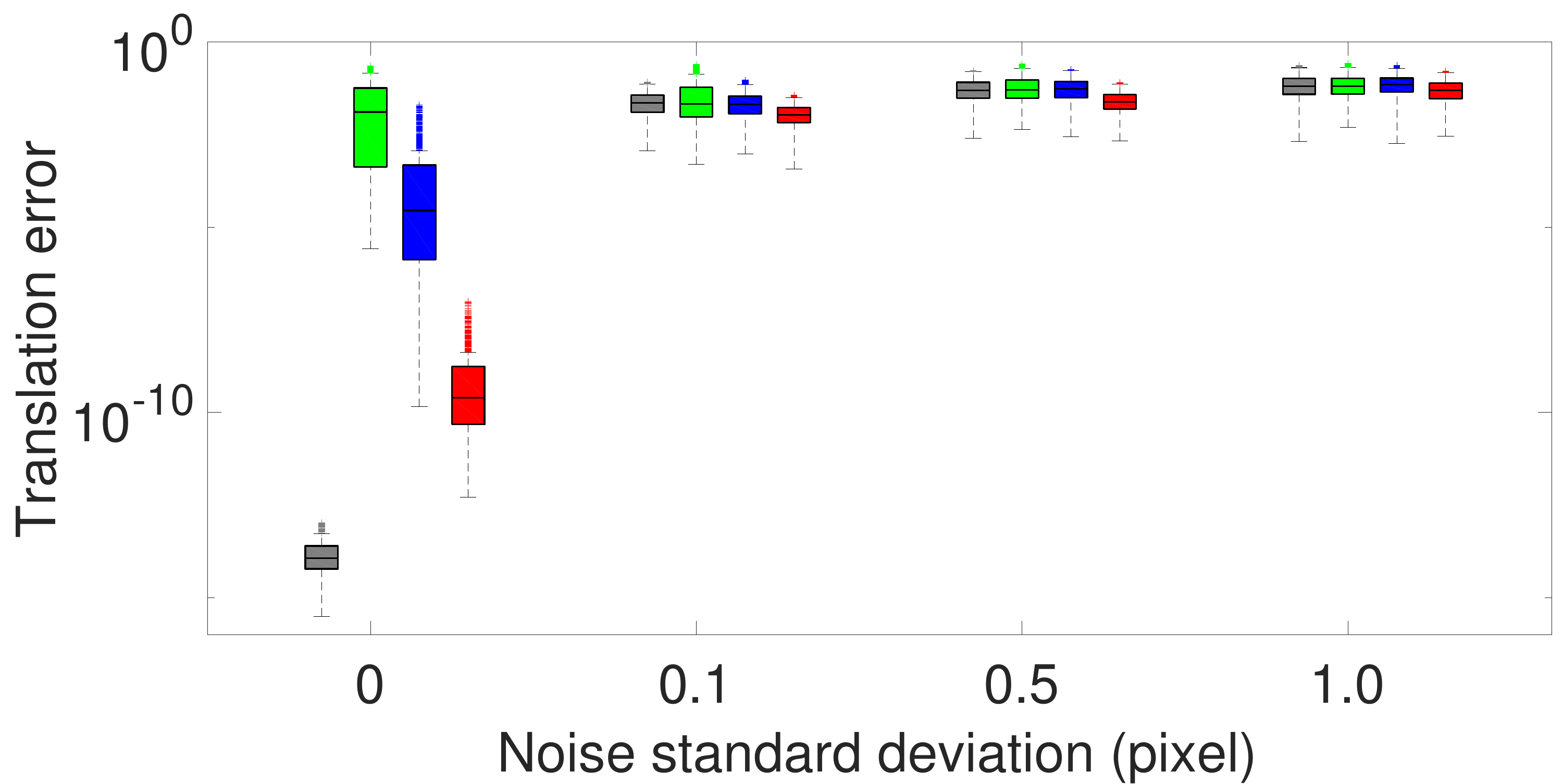}
		\caption{$\varepsilon_{\mathbf{t}}$ for random motion.}
	\end{subfigure}
	\hfill
	\begin{subfigure}{0.32\linewidth}
		\includegraphics[width=1.0\linewidth]{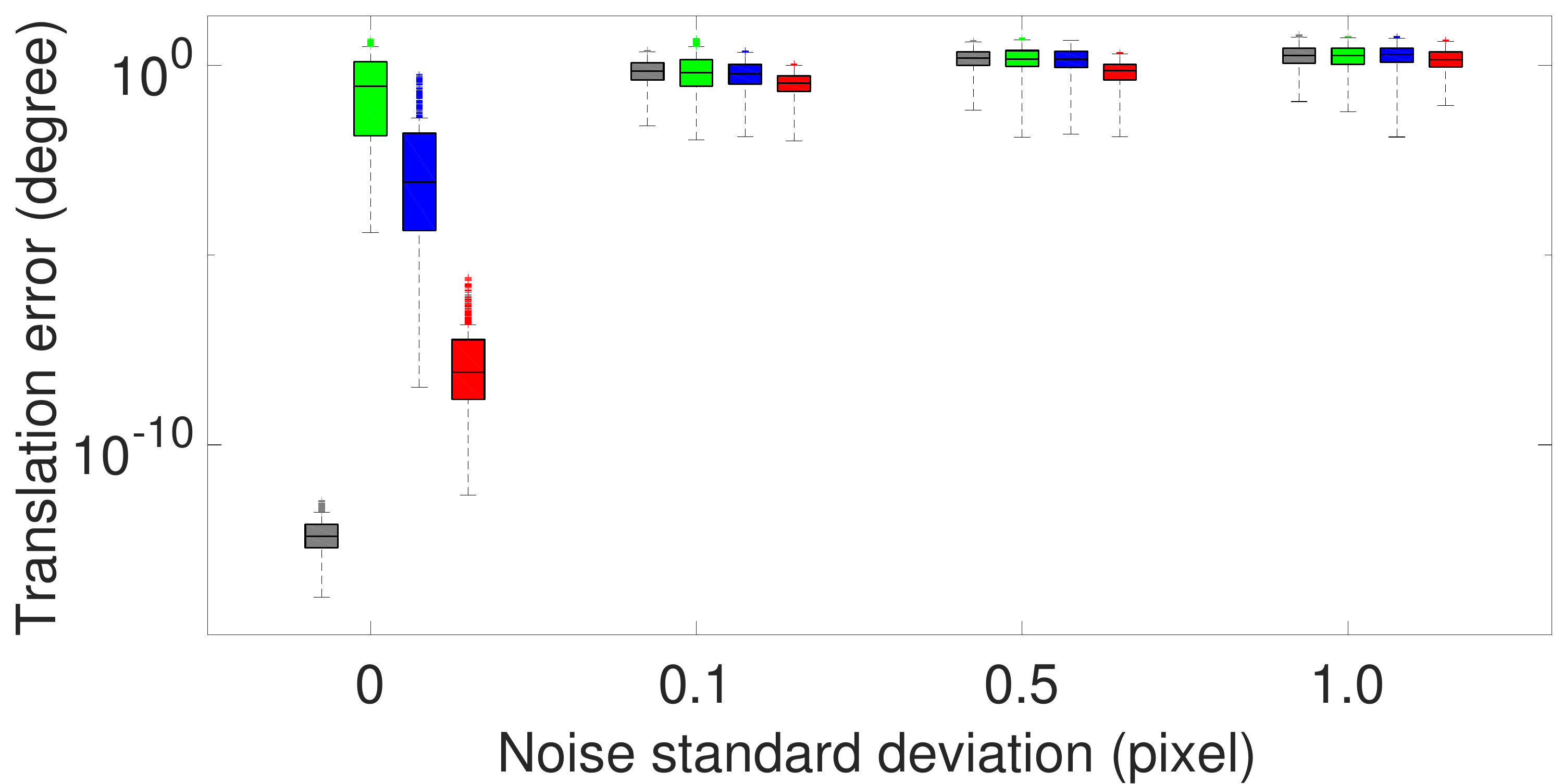}
		\caption{$\varepsilon_{\mathbf{t},\text{dir}}$ for random motion.}
	\end{subfigure}
	\caption{Rotation and translation error with varying image noise for the inter-camera case. We designed a simulated two-camera rig composed of two perspective cameras. The orientations of the two perspective cameras are roughly forward-facing with random perturbation. The resulting data is organized into three rows corresponding to forward, sideways, and random motions.}
	\label{fig:KITTI_2cameras_inter}
\end{figure*}

\subsubsection{Synthetic Experiments for Two-camera Rigs} 
In this scenario, we design a simulated two-camera rig composed of two perspective cameras. The orientations of the two perspective cameras are roughly forward-facing with random perturbation. This setting is practical for autonomous driving with two front-facing cameras~\cite{geiger2013vision}. The performance of the solvers is tested in inter-camera and intra-camera configurations. 

The baseline length between the two simulated cameras is fixed at $1$ meter. The multi-camera reference frame is defined at the center of the camera rig, with a translation distance of $3$ meters between two consecutive multi-camera reference frames. Each camera has a resolution of $640 \times 480$ pixels, and a focal length of $400$ pixels. The principal points of the cameras are set at the image center. The synthetic scene consists of a ground plane and $50$ randomly generated planes. These planes are created within a cube-shaped region with dimensions of $[-5, 5] \times [-5, 5] \times [10, 20]$ meters, defined in the corresponding axes of the multi-camera reference frame. Within the synthetic data, we randomly select $50$ PCs belonging to the ground plane, as well as a PC from each randomly generated plane. Therefore, a total of $100$ PCs are generated randomly in the synthetic scene. 

The synthetic experiment consists of a series of $1000$ trials, wherein each trial entails the random generation of $100$ PCs. PCs are also randomly selected for the different solvers. The error evaluation is focused on determining the best relative pose, which yields the maximum inlier count within the RANSAC scheme. The RANSAC scheme also allows the selection of the most suitable candidate from multiple solutions. Rotation and translation errors are assessed using the median of errors. To induce different types of motion, the translation direction between two multi-camera reference frames is intentionally chosen to create forward, sideways, or random motions. For each motion, the second view is subjected to a random rotation, which is sequentially performed around three axes with the rotation angles ranging from $-10^\circ$ to $10^\circ$.
\begin{figure*}[t]
	\centering
	\includegraphics[width=0.50\linewidth]{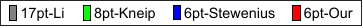}\\
	\centering
	\begin{subfigure}{0.32\linewidth}
		\includegraphics[width=1.0\linewidth]{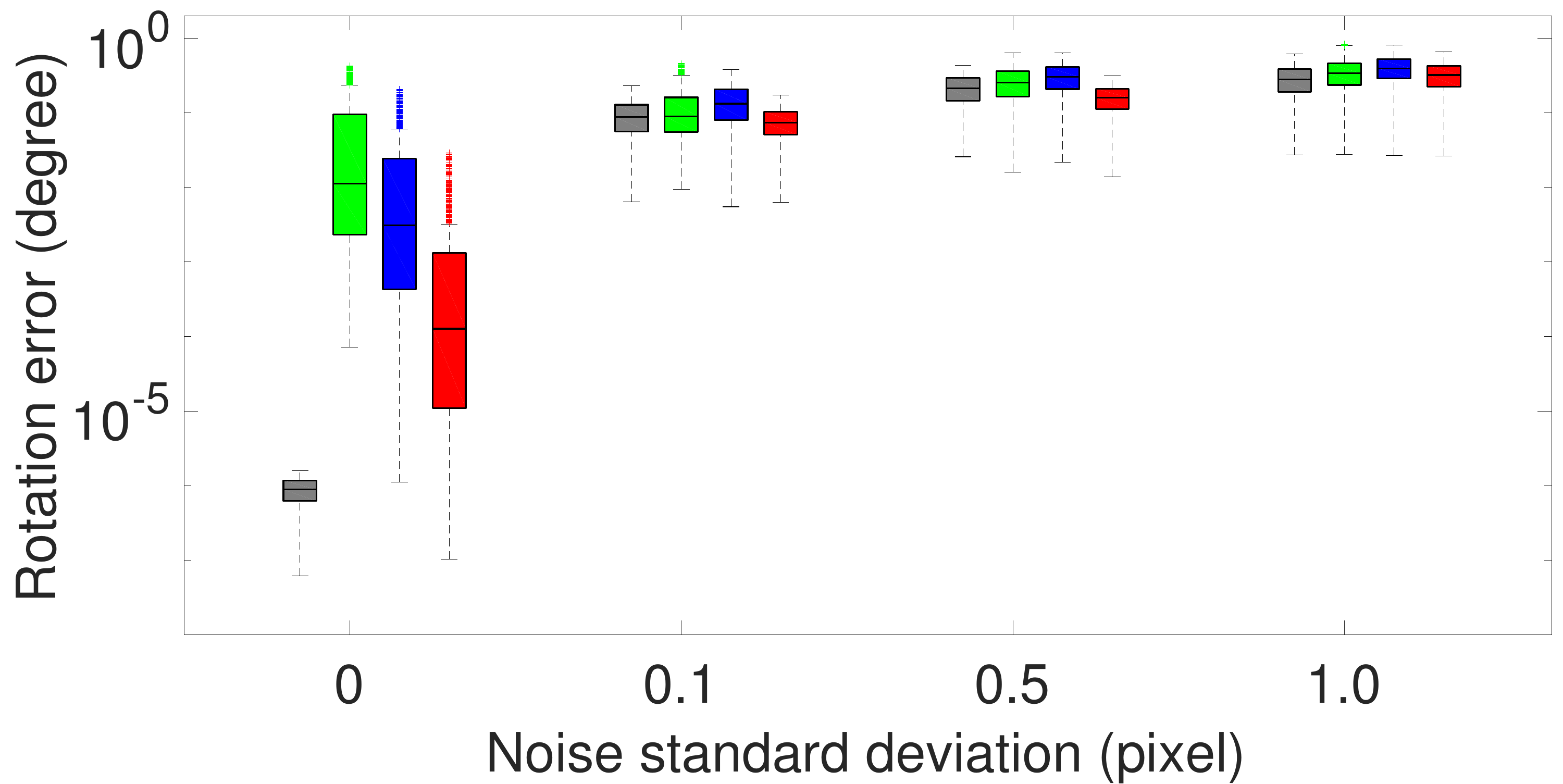}
		\caption{${\varepsilon_{\mathbf{R}}}$ for forward motion.}
	\end{subfigure}
	\hfill
	\begin{subfigure}{0.32\linewidth}
		\includegraphics[width=1.0\linewidth]{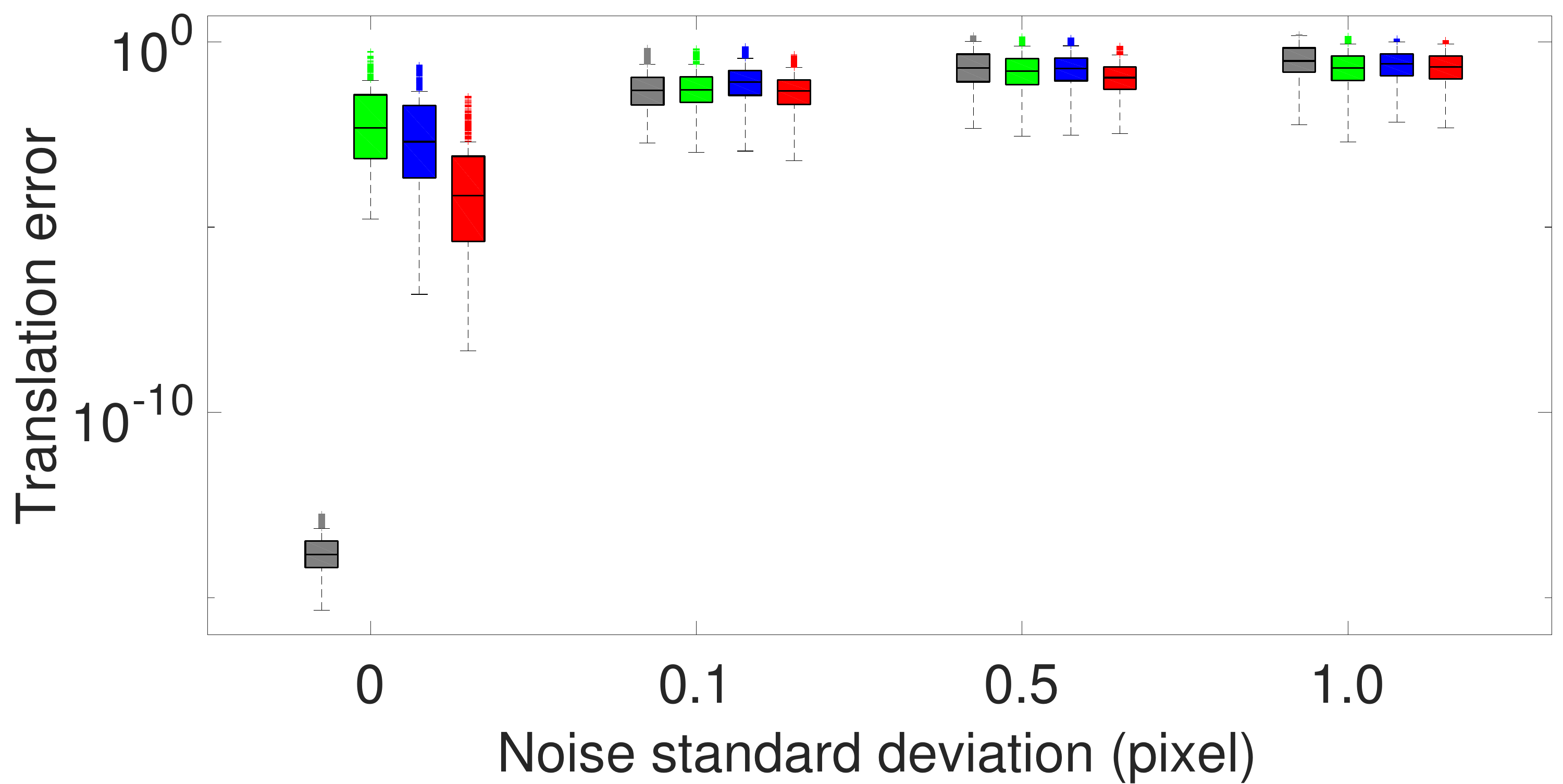}
		\caption{$\varepsilon_{\mathbf{t}}$ for forward motion.}
	\end{subfigure}
	\hfill
	\begin{subfigure}{0.32\linewidth}
		\includegraphics[width=1.0\linewidth]{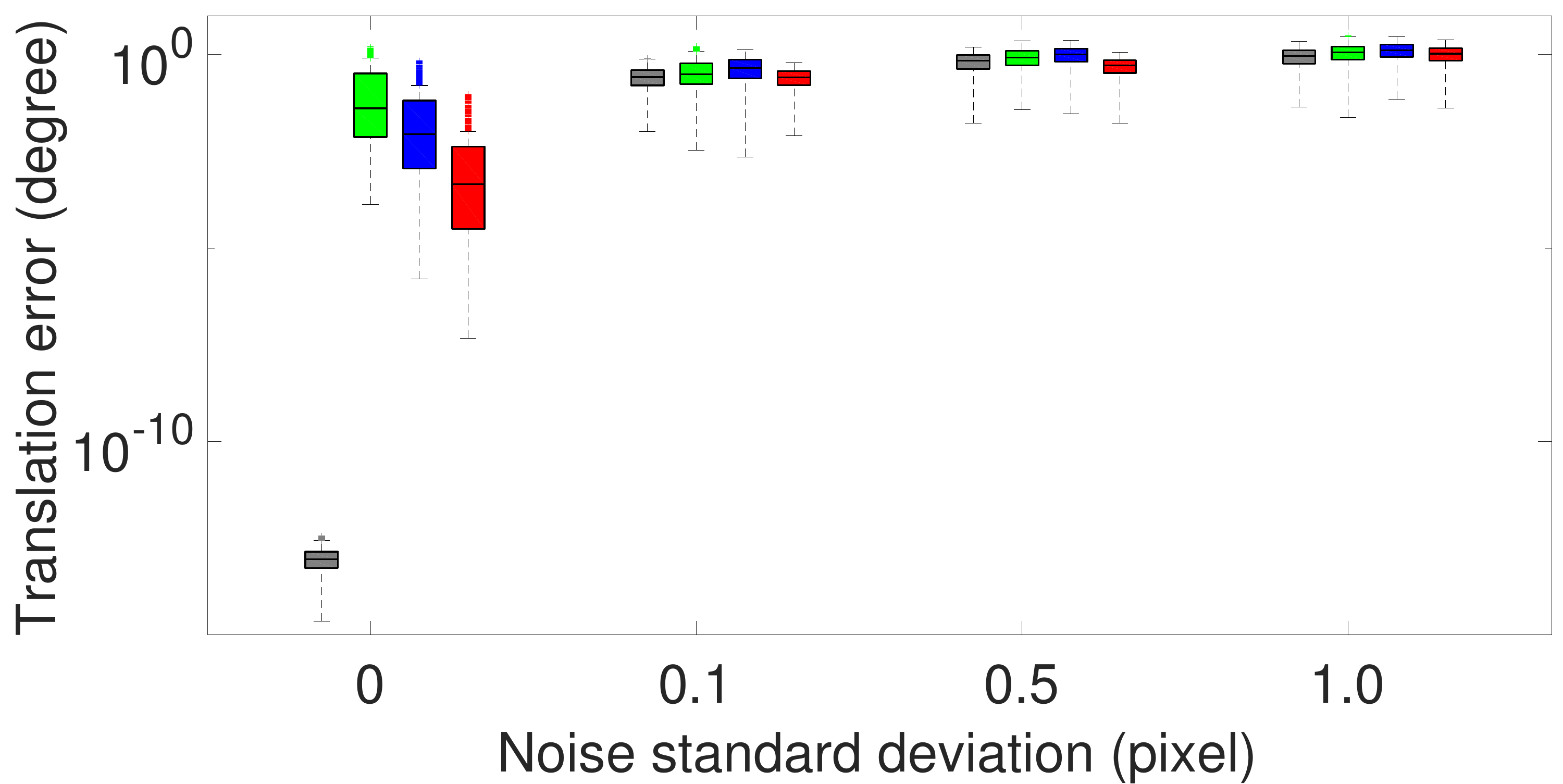}
		\caption{$\varepsilon_{\mathbf{t},\text{dir}}$ for forward motion.}
	\end{subfigure}
	\begin{subfigure}{0.32\linewidth}
		\includegraphics[width=1.0\linewidth]{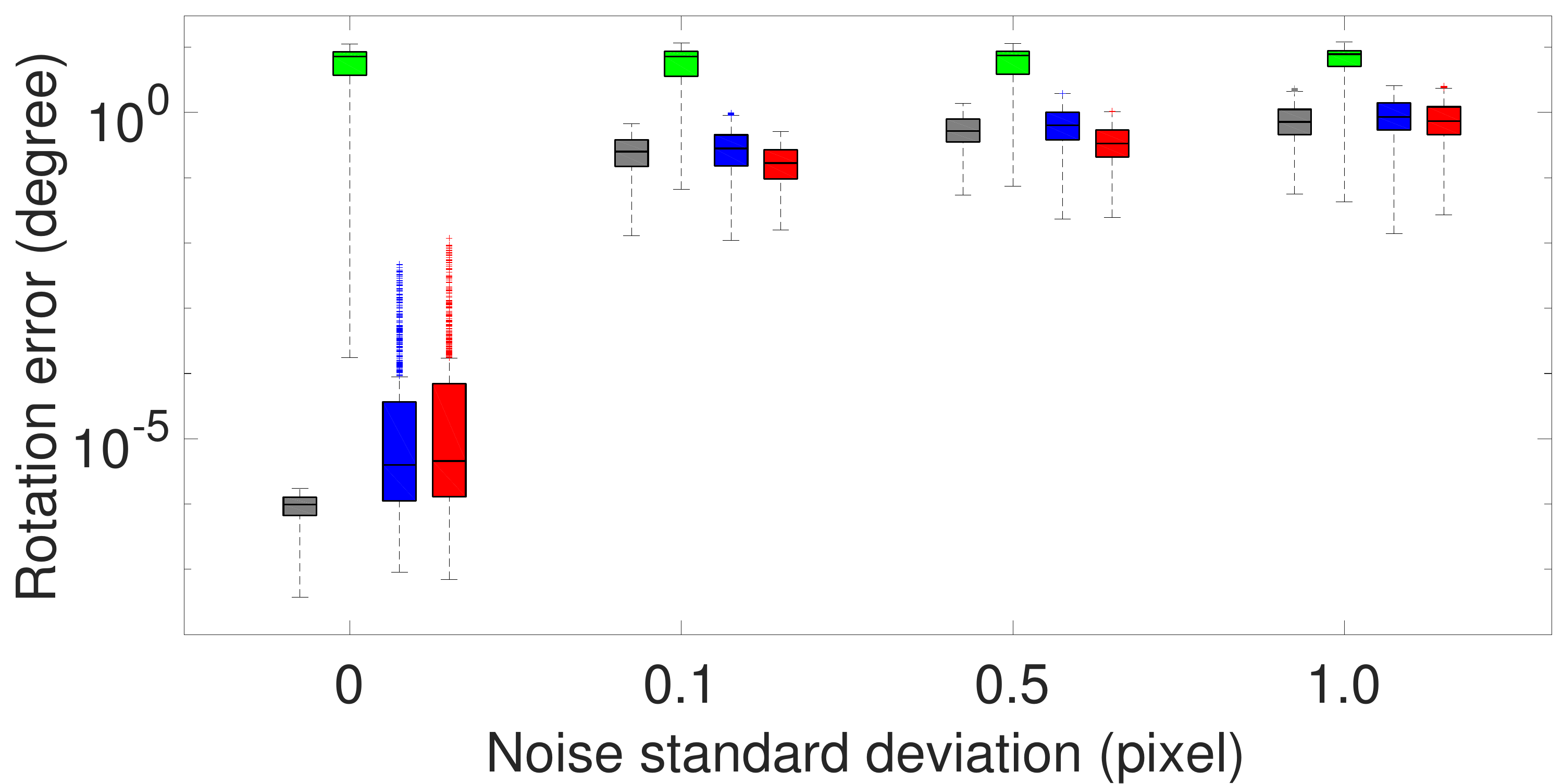}
		\caption{${\varepsilon_{\mathbf{R}}}$ for sideways motion.}
	\end{subfigure}
	\hfill
	\begin{subfigure}{0.32\linewidth}
		\includegraphics[width=1.0\linewidth]{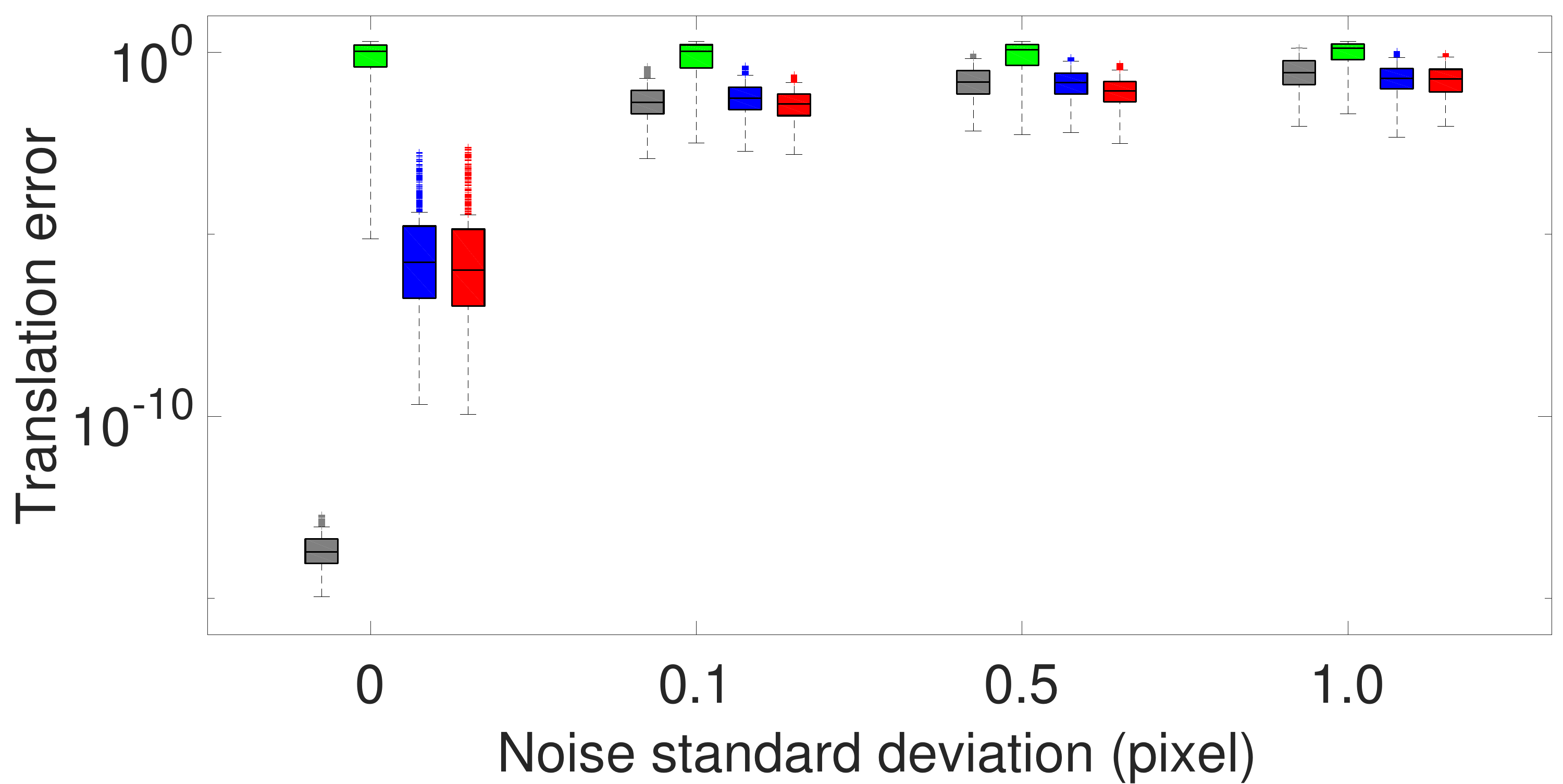}
		\caption{$\varepsilon_{\mathbf{t}}$ for sideways motion.}
	\end{subfigure}
	\hfill
	\begin{subfigure}{0.32\linewidth}
		\includegraphics[width=1.0\linewidth]{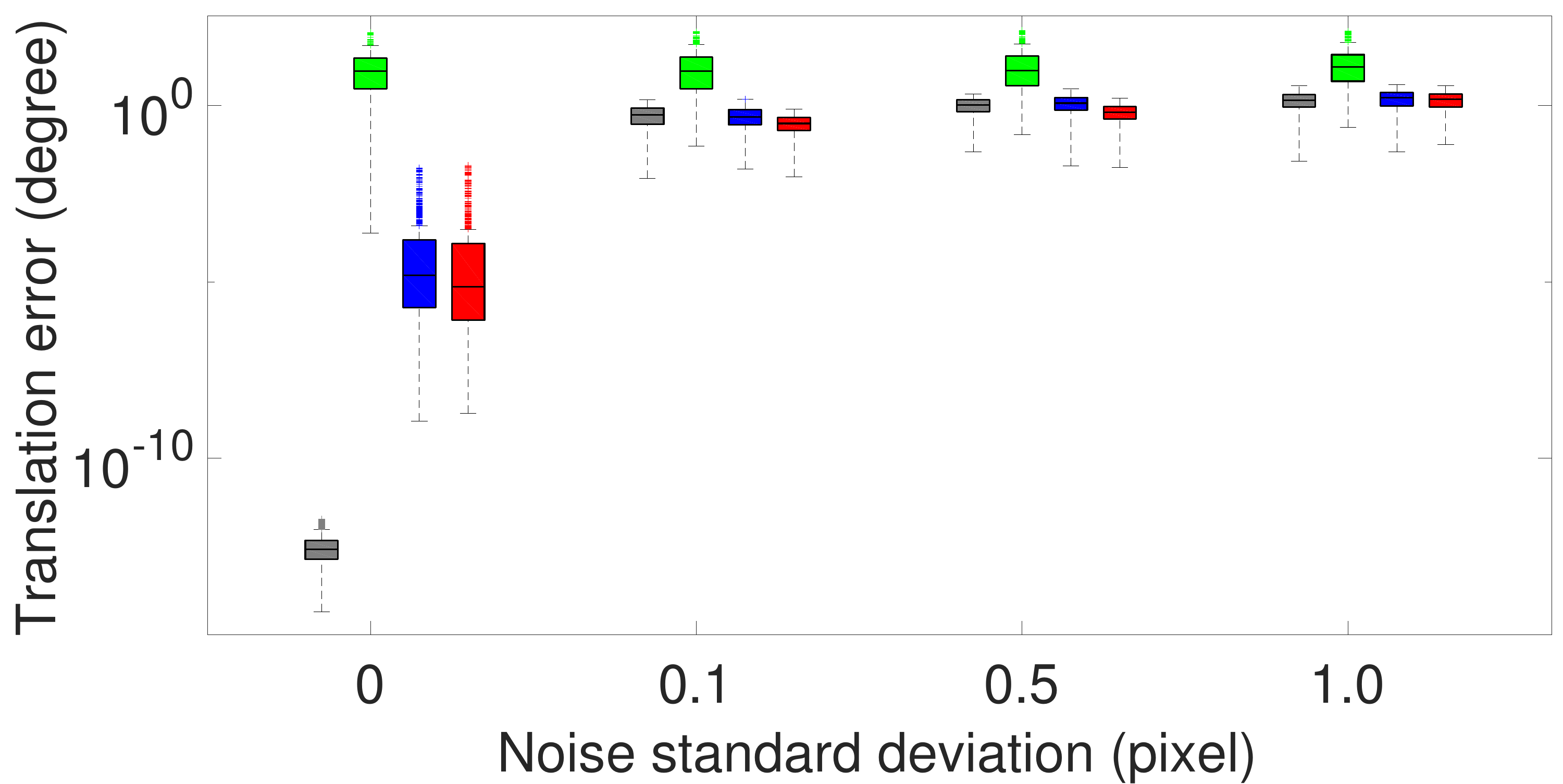}
		\caption{$\varepsilon_{\mathbf{t},\text{dir}}$ for sideways motion.}
	\end{subfigure}
	\begin{subfigure}{0.32\linewidth}
		\includegraphics[width=1.0\linewidth]{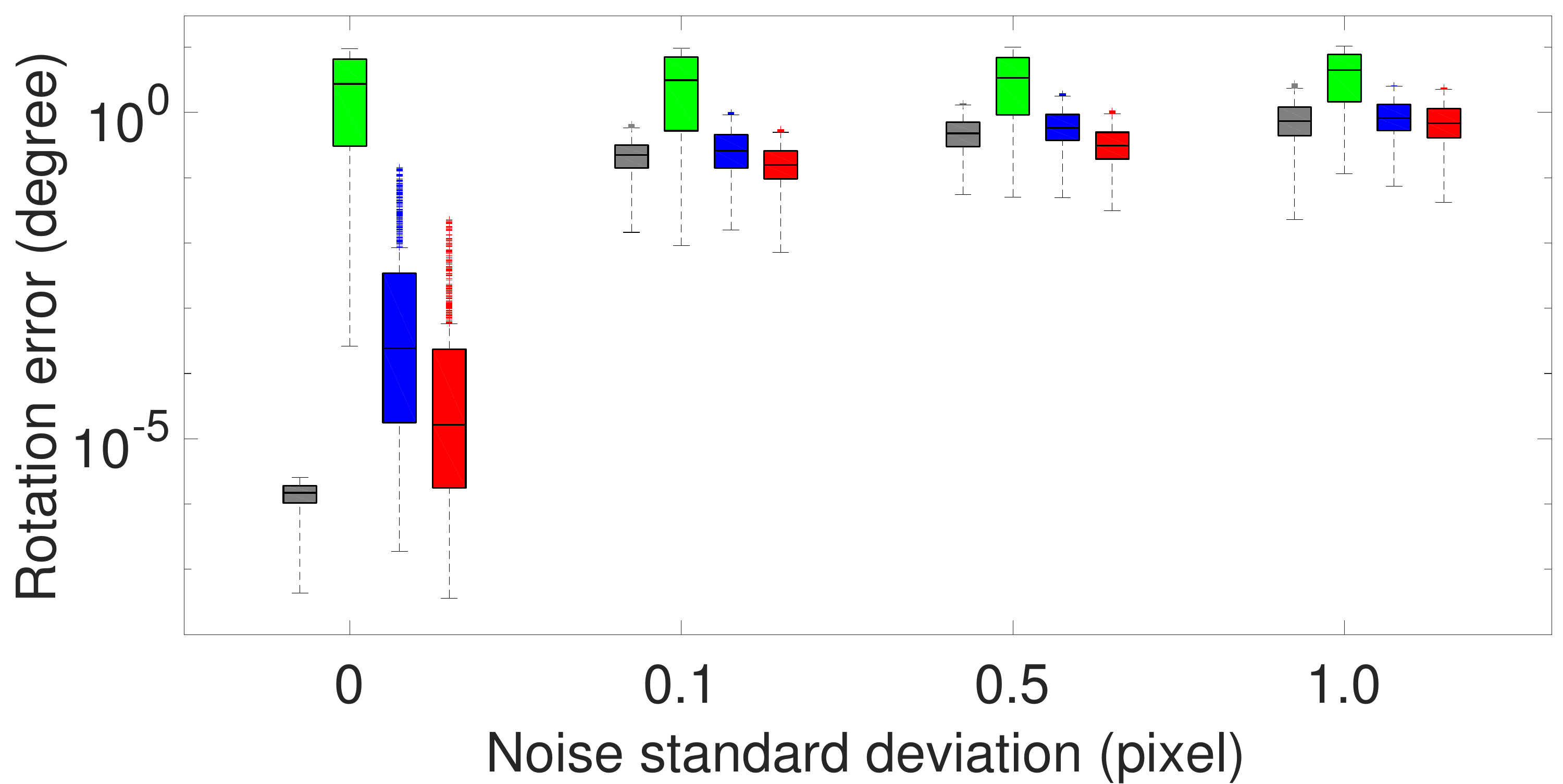}
		\caption{${\varepsilon_{\mathbf{R}}}$ for random motion.}
	\end{subfigure}
	\hfill
	\begin{subfigure}{0.32\linewidth}
		\includegraphics[width=1.0\linewidth]{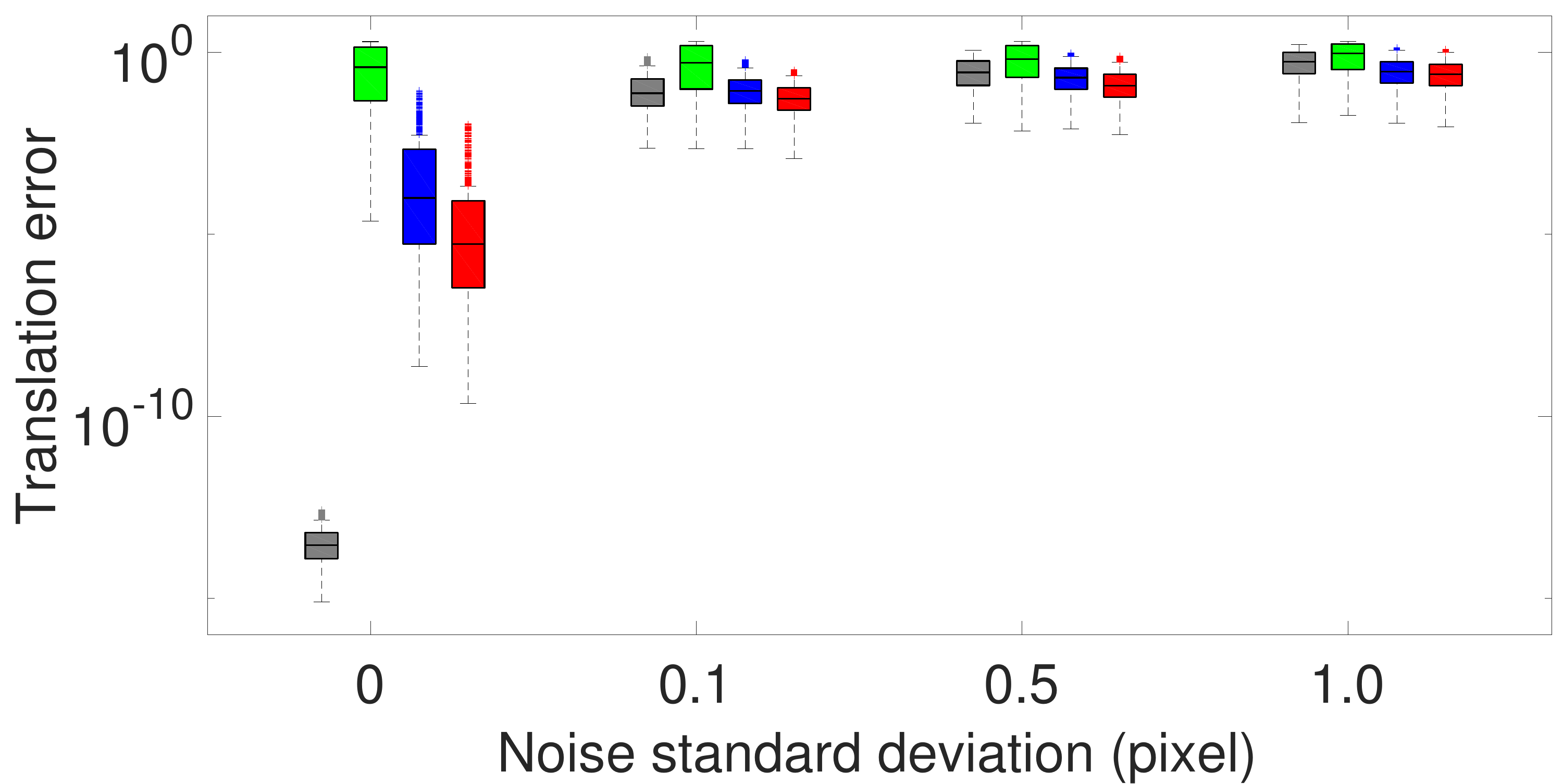}
		\caption{$\varepsilon_{\mathbf{t}}$ for random motion.}
	\end{subfigure}
	\hfill
	\begin{subfigure}{0.32\linewidth}
		\includegraphics[width=1.0\linewidth]{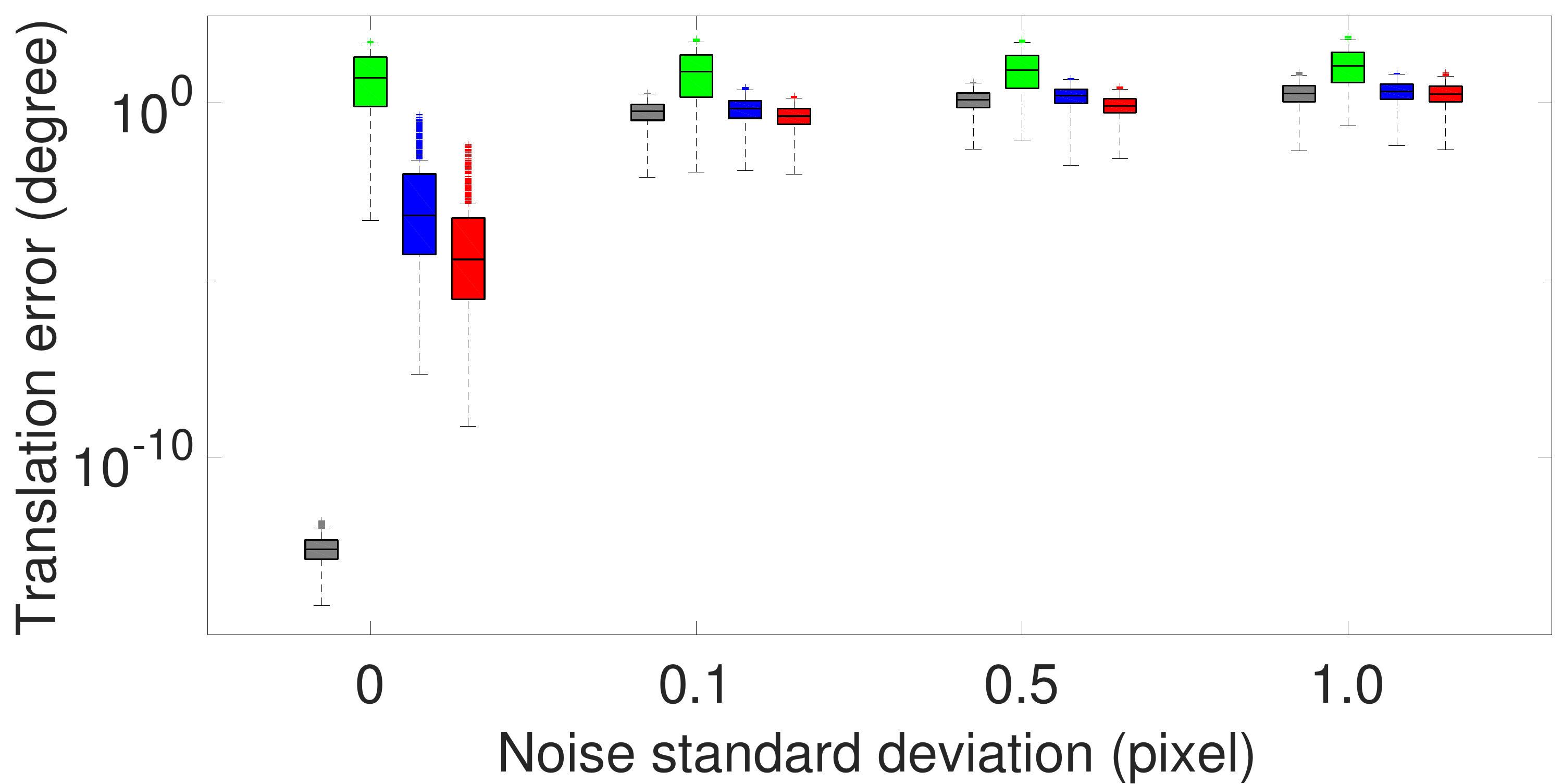}
		\caption{$\varepsilon_{\mathbf{t},\text{dir}}$ for random motion.}
	\end{subfigure}
	\caption{Rotation and translation error with varying image noise for the intra-camera case. We designed a simulated two-camera rig composed of two perspective cameras. The orientations of the two perspective cameras are roughly forward-facing with random perturbation. The resulting data is organized into three rows corresponding to forward, sideways, and random motions.}
	\label{fig:KITTI_2cameras_intra}
\end{figure*}

\cref{fig:KITTI_2cameras_inter} illustrates the performance comparison of different solvers with respect to image noise for the inter-camera case. A good method should have small median and variance for  ${\varepsilon_{\mathbf{R}}}$, ${\varepsilon_{\mathbf{t}}}$, and $\varepsilon_{\mathbf{t},\text{dir}}$. The following observations are made: (1)~The proposed \texttt{6pt-Our-inter} solver demonstrates superior results compared to the comparative solvers. (2)~The iterative optimization utilized  in \texttt{8pt-Kneip} is prone to convergence to local minima. This solver performs well in addressing forward and random motions. However, it does not perform well for the sideways motion. (3)~The linear solver \texttt{17pt-Li} with fewer calculations have less round-off error than the proposed \texttt{6pt-Our-inter} solver in noise-free cases. However, our method has better accuracy than the linear solver with the influence of image noise. This is also consistent with real-world data experiments.

\cref{fig:KITTI_2cameras_intra} illustrates the performance comparison of different solvers with respect to image noise for the intra-camera case. The following observations are made: (1) Compared to the results in \cref{fig:KITTI_2cameras_inter}, the solvers using intra-camera correspondences generally perform worse than inter-camera correspondences, particularly in terms of recovering the metric scale of translation. (2) The proposed \texttt{6pt-Our-intra} solver has better performance than the comparative solvers. (3) The \texttt{8pt-Kneip} solver demonstrates satisfactory performance in the scenario of forward motion within multi-camera systems, but it displays diminished effectiveness in scenarios involving sideways and random motions.

\subsubsection{\label{sec:SynexperimentsGene}Synthetic Experiments for Generalized Camera}
\begin{table}[tbp]
	\caption{The correspondence selection for different solvers. The pair $(i, i')$ denotes a PC that is observed in the $i$-th camera in the first view and the $i'$-th camera in the second view. The multiplier after a pair means the number of this match type.}
	\begin{center}
		\setlength{\tabcolsep}{1.0mm}{
			\scalebox{0.88}{
				\begin{tabular}{lccc}
					\toprule
					{17pt-Li}~\cite{li2008linear} & {8pt-Kneip}~\cite{kneip2014efficient} &  {6pt-Stew{\'e}nius}~\cite{stewenius2005solutions} &  6pt-Our-generic \\
					\midrule
					(1,2)*3   & (1,2)*2   & (1,2)*1   & (1,2)*1   \\
					(3,4)*3   & (3,4)*2   & (3,4)*1   & (3,4)*1   \\
					(5,6)*3   & (5,6)*1   & (5,6)*1   & (5,6)*1   \\
					(7,8)*3   & (7,8)*1   & (7,8)*1   & (7,8)*1   \\
					(9,10)*3  & (9,10)*1  & (9,10)*1  & (9,10)*1  \\
					(11,12)*2 & (11,12)*1 & (11,12)*1 & (11,12)*1 \\
					\bottomrule
		\end{tabular}}}
	\end{center}
	\label{tab:correspondece_selection}
\end{table}
In this scenario, we design a simulated generalized camera composed of 12 omnidirectional cameras. The extrinsic parameters, including orientation and position, are totally random. Some of the comparison solvers use more PCs than the proposed solvers. The strategy of correspondence selection has a significant influence on their performance. We design a rule to select matches for different cases to guarantee fairness. Take the generic-camera configuration as an example. The match types are $(1,2)$, $(3, 4)$, $(5, 6)$, $(7, 8)$, $(9, 10)$, and $(11, 12)$ for the proposed generic solver. Recall that the pair $(i, i')$ denotes a PC that is observed in the $i$-th camera in the first view and the $i'$-th camera in the second view. Solvers \texttt{17pt-Li}, \texttt{8pt-Kneip}, and \texttt{6pt-Stew{\'e}nius} cyclically build matches according to the order of $(1,2)$, $(3, 4)$, $(5, 6)$, $(7, 8)$, $(9, 10)$, and $(11, 12)$. \cref{tab:correspondece_selection} shows the correspondence selection for different solvers.

\subsection{\label{sec:KITTIexperiments}Experiments on KITTI Dataset}
\begin{figure}[tbp]
	\centering
	\begin{subfigure}{0.6\linewidth}
		\includegraphics[width=1.0\linewidth]{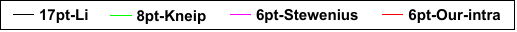}
	\end{subfigure}
	\begin{subfigure}{0.48\linewidth}
		\includegraphics[width=1.0\linewidth]{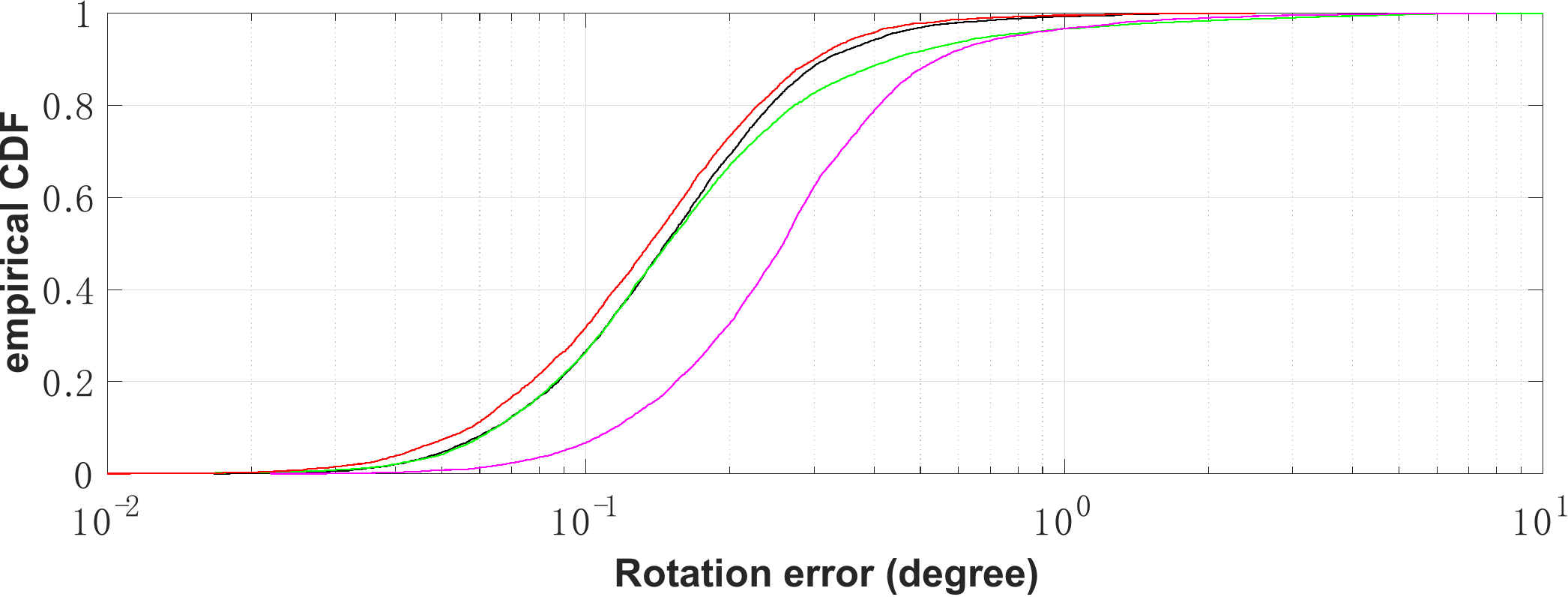}
		\caption{Rotation error.}
	\end{subfigure}
	\begin{subfigure}{0.48\linewidth}
		\includegraphics[width=1.0\linewidth]{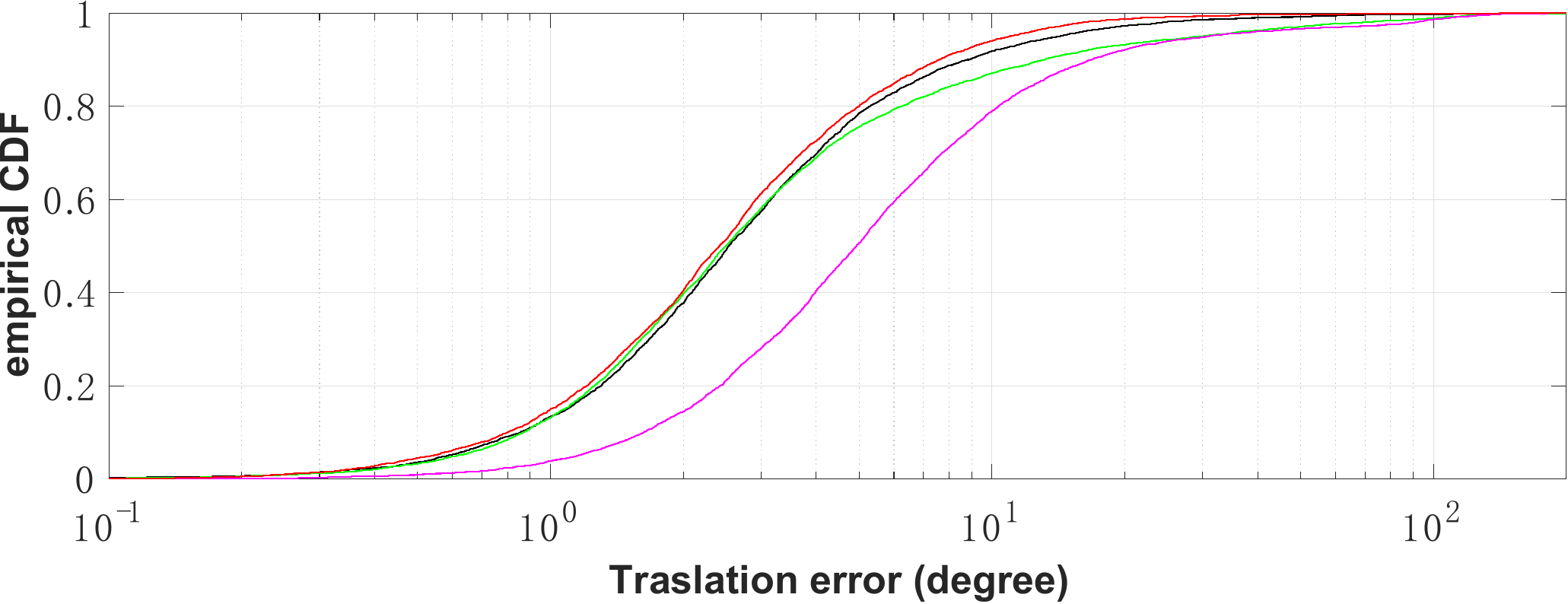}
		\caption{Translation error.}
	\end{subfigure}
	\caption{Empirical cumulative error distributions for \texttt{KITTI} sequence 00.}
	\label{fig:RTCDF_KITTI}
\end{figure}

\cref{fig:RTCDF_KITTI} presents the empirical cumulative error distributions for sequence 00 in \texttt{KITTI} dataset. These distributions are derived from the same underlying values utilized to construct Table~2 in the paper. The results showcase that the proposed \texttt{6pt-Our-intra} solver outperforms state-of-the-art solvers, displaying the most favorable overall performance. In addition, we have compared the suggested experiments with the solvers \texttt{6pt-Ventura}~\cite{ventura2015efficient} and \texttt{4pt-Sweeney}~\cite{sweeney2014solving}, which utilize a prior for relative pose estimation. The median rotation and translation error on Sequence 00 of the \texttt{KITTI} dataset: \texttt{6pt-Ventura} ($0.141^\circ$, $2.458^\circ$), \texttt{4pt-Sweeney} ($0.056^\circ$, $2.289^\circ$). Our solver outperforms \texttt{6pt-Ventura} in terms of accuracy. Since \texttt{4pt-Sweeney} assumes known vertical direction as a prior, it achieves the highest accuracy. Given the proposed solvers do not exploit any motion prior, this comparative result does not actually harm our contribution.

\subsection{\label{sec:nuScenesexperiments}Experiments on nuScenes Dataset}
The performance evaluation of the solvers extends to the \texttt{nuScenes} dataset~\cite{Caesar_2020_CVPR}. This dataset comprises consecutive keyframes captured from a multi-camera system consisting of six cameras, providing a comprehensive 360-degree field of view. For evaluation purposes, all keyframes from Part 1 are utilized, totaling 3,376 images. The ground truth is established through a lidar map-based localization scheme. Similar to the experiments conducted on the \texttt{KITTI} dataset, the SIFT detector~\cite{lowe2004distinctive} is employed to establish PCs between consecutive views in the six cameras. To deal with outlier matches in the feature correspondences, all the solvers are integrated into the RANSAC scheme.
\begin{table}[tbp]
	\caption{Rotation and translation error on \texttt{nuScenes} dataset (unit: degree).}
	\begin{center}
		\setlength{\tabcolsep}{1.0mm}{
			\scalebox{1.00}{
				\begin{tabular}{c||c|c|c|c}
					\hline
					\multirow{2}{*}{{Part}} &  
					{17pt-Li}\small{~\cite{li2008linear}} &  {8pt-Kneip}\small{~\cite{kneip2014efficient}} &  {6pt-Stew.}\small{~\cite{stewenius2005solutions}}&  {6pt-Our-intra} \\
					\cline{2-5}
					& ${\varepsilon _{\bf{R}}}$\quad\ $\varepsilon_{\mathbf{t},\text{dir}}$      &  ${\varepsilon _{\bf{R}}}$\quad\ $\varepsilon_{\mathbf{t},\text{dir}}$      &   ${\varepsilon _{\bf{R}}}$\quad\ $\varepsilon_{\mathbf{t},\text{dir}}$     &   ${\varepsilon _{\bf{R}}}$\quad\ $\varepsilon_{\mathbf{t},\text{dir}}$  \\
					\hline
					{01}&  0.183 \ 2.826 &  0.175 \  2.732 & 0.207 \  2.898  & \textbf{0.155} \ \textbf{2.519} \\
					\hline						
		\end{tabular}}}
	\end{center}
	\label{tab:RTErrror_nuScenes_generalized}
\end{table}

\cref{tab:RTErrror_nuScenes_generalized} illustrates the rotation and translation error of the \texttt{6pt-Our-intra} solver on the Part1 of \texttt{nuScenes} dataset. We use median error to evaluate the performance of solvers. The experiment results highlight that the proposed \texttt{6pt-Our-intra} solver achieves superior performance compared to other solvers. In addition, compared to experiments conducted on the \texttt{KITTI} dataset, this experiment provides further evidence supporting the direct applicability of the \texttt{6pt-Our-intra} method for relative pose estimation in systems equipped with more cameras.

\section{Limitations}
The vanilla RANSAC is used due to its simplicity and few parameters. SOTA RANSAC variants are useful to push the envelope of performance for a method, but it is hard to guarantee fairness when comparing different solvers. Though the performance of the vanilla RANSAC is not as good as SOTA RANSAC variants, it is sufficient to provide a fair and clear benchmark for all minimal solvers.

\end{sloppypar}


\end{document}